\documentclass{article}

\PassOptionsToPackage{numbers, compress}{natbib}

\usepackage[preprint]{neurips_2025}

\usepackage[utf8]{inputenc} 
\usepackage[T1]{fontenc}    
\usepackage{hyperref}       
\usepackage{url}            
\usepackage{booktabs}       
\usepackage{amsfonts}       
\usepackage{nicefrac}       
\usepackage{microtype}      
\usepackage{xcolor}         
\usepackage{array}

\usepackage{amsthm}

\usepackage{cite}
\usepackage{enumerate}
\usepackage{graphics} 
\usepackage{epsfig} 
\usepackage{libertine}
\usepackage{mathptmx} 
\usepackage{times} 
\usepackage{amsmath} 
\usepackage{dutchcal}
\usepackage{makecell}

\usepackage{amssymb}  
\usepackage{mathtools}
\usepackage{bbm}
\usepackage{mathrsfs}

\usepackage{algorithm} 
\usepackage{algpseudocode}
\usepackage{listings}
\usepackage{hyperref} 
\usepackage{booktabs} 
\usepackage{xcolor} 
\usepackage{graphicx}
\usepackage{subfigure}
\usepackage{tabularx} 
\usepackage{bbm}
\usepackage{enumitem}
\usepackage{amsmath,amssymb,amsthm}
\usepackage[export]{adjustbox}
\usepackage{changes}
\usepackage{multirow}
\usepackage{cleveref}
\usepackage[table]{xcolor}
\usepackage{wrapfig}
\usepackage[most]{tcolorbox}

\usepackage[table]{xcolor} 
\usepackage{colortbl} 
\usepackage{ragged2e} 
\usepackage{soul}     

\definecolor{mylightgray}{gray}{0.92}
\definecolor{mylightgreen}{RGB}{200, 255, 200} 
\definecolor{mylightred}{RGB}{255, 200, 200}   
\definecolor{mytextgreen}{RGB}{0, 150, 0}   
\definecolor{mytextred}{RGB}{200, 0, 0}     

\definecolor{mylightgray}{gray}{0.92}
\definecolor{mylightgreen}{RGB}{200, 255, 200} 
\definecolor{mylightred}{RGB}{255, 200, 200}   
\definecolor{mytextgreen}{RGB}{0, 150, 0}   
\definecolor{mytextred}{RGB}{200, 0, 0}     
\definecolor{mycustomlightblue}{HTML}{D9EAF7} 
\definecolor{mycustomlightred}{HTML}{FEE7E9}   

\newcommand{\greentext}[1]{{\color{mytextgreen}#1}}
\newcommand{\redtext}[1]{{\color{mytextred}#1}}
\newcommand{\libertinefont}{\fontfamily{LinuxLibertineT-OsF}\selectfont}

\newcommand{\boxredhighlight}[1]{%
  \tcbox[
    on line,
    size=fbox,
    arc=1.5pt,
    colback=mycustomlightred,
    colframe=mycustomlightred,
    boxrule=0pt,
    boxsep=1pt,
    left=0.5pt,
    right=0.5pt,
    top=0.0pt,
    bottom=0.0pt,
    nobeforeafter,
    tcbox raise base,
    fontupper=\libertinefont
  ]{#1}%
}

\title{Distribution Preference Optimization: A Fine-grained Perspective for LLM Unlearning}

\author{%
  Kai Qin\textsuperscript{1},
  Jiaqi Wu\textsuperscript{1},
  Jianxiang He\textsuperscript{2},
  Haoyuan Sun\textsuperscript{1},
  Yifei Zhao\textsuperscript{1}, 
  Xu Wang\textsuperscript{3}, \\
  \textbf{Bin Liang\textsuperscript{4},
  Yongzhe Chang\textsuperscript{1},
  Cheng Li\textsuperscript{3},
  Tiantian Zhang\textsuperscript{1},
  Houde Liu\textsuperscript{1}} \\
  \textsuperscript{1}Tsinghua University \quad
  \textsuperscript{2}The Hong Kong University of Science and Technology \\
  \textsuperscript{3}Jianghuai Advanced Technology Center \quad
  \textsuperscript{4}University of Technology Sydney
}

\begin{document}

\maketitle

\begin{abstract}
As Large Language Models (LLMs) demonstrate remarkable capabilities learned from vast corpora, concerns regarding data privacy and safety are receiving increasing attention. LLM unlearning, which aims to remove the influence of specific data while preserving overall model utility, is becoming an important research area. One of the mainstream unlearning classes is optimization-based methods, which achieve forgetting directly through fine-tuning, exemplified by Negative Preference Optimization (NPO). However, NPO's effectiveness is limited by its inherent lack of explicit positive preference signals. Attempts to introduce such signals by constructing preferred responses often necessitate domain-specific knowledge or well-designed prompts, fundamentally restricting their generalizability. In this paper, we shift the focus to the distribution-level, directly targeting the next-token probability distribution instead of entire responses, and derive a novel unlearning algorithm termed \textbf{Di}stribution \textbf{P}reference \textbf{O}ptimization (DiPO). We show that the requisite preference distribution pairs for DiPO, which are distributions over the model's output tokens, can be constructed by selectively amplifying or suppressing the model's high-confidence output logits, thereby effectively overcoming NPO's limitations. We theoretically prove the consistency of DiPO's loss function with the desired unlearning direction. Extensive experiments demonstrate that DiPO achieves a strong trade-off between model utility and forget quality. Notably, DiPO attains the highest forget quality on the TOFU benchmark, and maintains leading scalability and sustainability in utility preservation on the MUSE benchmark.
\end{abstract}

\section{Introduction}
\label{s:intro}
The increasing capabilities and widespread application of Large Language Models (LLMs) trained on massive corpora are accompanied by significant ethical and safety challenges. These include the risk of generating biased or offensive content \citep{yu2023unlearning, liu2022continual, eldan2023s}, concerns over data privacy and copyright \citep{jang2022knowledge, wu2023depn, liu2024towards}, and potential misuse \citep{barrett2023identifying}. Regulatory frameworks \citep{GDPR2016, CCPA_AB375_2018} , with their ``Right to be Forgotten'' provisions, impose legal obligations to remove user data. The need to effectively remove the influence of specific information from trained LLMs, particularly to prevent its leakage, has motivated research into \emph{LLM unlearning}. This area focuses on developing methods to achieve such selective erasure without compromising the model's overall utility \citep{yao2024large, zhang2024negative}.

Among existing approaches, \emph{optimization-based methods}, which directly fine-tune model parameters to induce forgetting, represent a mainstream paradigm. Gradient Ascent (GA) \citep{jang2022knowledge, yao2024large}, for example, maximizes the token prediction loss on the forget set to achieve forgetting. Yet, unbounded maximization often leads to model instability and performance degradation. Negative Preference Optimization (NPO) \citep{zhang2024negative} is proposed to mitigate this issue by employing a bounded forgetting loss modified from Direct Preference Optimization (DPO) \citep{rafailov2023direct}.

However, the lack of positive preference signals limits the effectiveness of NPO. Attempts to reintroduce such signals face significant challenges: using template-based alternative responses (e.g. I don't know) often induces \emph{catastrophic forgetting}, while generating higher-quality alternatives typically requires domain-specific knowledge and thus limits its applicability and efficiency.
We posit that \emph{this challenge fundamentally stems from the nature of the response-level}: the vast and unstructured space of possible responses makes the construction of suitable preferred responses inherently difficult.

In this paper, we propose shifting the focus to the \textbf{distribution-level}, targeting the next-token probability distribution directly, \emph{as the model's vocabulary table provides the complete and crucially, finite set of all possible alternative tokens}. 
Drawing from this perspective and defining the distribution-level immediate reward, we derive a novel algorithm termed \textbf{Di}stribution \textbf{P}reference \textbf{O}ptimization (DiPO). We show that the requisite preference distribution pairs can be intrinsically constructed via logit modulation, enabling effective unlearning without auxiliary components. Intuitively, the DiPO loss function effectively encourages an increase in the relative gap between the Sequence KL (SeqKL) divergence from the current distribution $\pi_{\theta}$ to prefered distribution $\pi_w$ and that to disprefered distribution $\pi_l$ (i.e. maximizing $D_{SeqKL}(x, y; \pi_l||\pi_{\theta}) - D_{SeqKL}(x, y; \pi_w||\pi_{\theta})$), incorporating a dynamic, per-sample offset. Further theoretical analysis of its gradient confirms that DiPO explicitly updates to move closer to $\pi_w$ and further away from $\pi_l$. 

As shown in \Cref{tab:sample_table}, DiPO consistently generates appropriate responses for both forget and retain queries. We conduct comprehensive experiments across various scenarios, including TOFU\citep{maini2024tofu} and MUSE\citep{shi2025muse}. On the TOFU benchmark, DiPO achieves new state-of-the-art performance, attaining a remarkable forget quality score of 0.86 for TOFU-10\%—nearly doubling the most competitive baseline's performance (0.45).  Furthermore, DiPO maintains leading performance on the MUSE benchmark, demonstrating superior scalability and sustainable utility preservation.
Our main contributions are as follows:



\begin{enumerate}
    \item We introduce distribution-level unlearning, directly optimizing the next-token probability distribution, which bypasses the explicit construction of preferred responses.
    \item We derive a novel unlearning algorithm termed \textbf{Di}stribution \textbf{P}reference \textbf{O}ptimization (DiPO), and theoretically prove the consistency of DiPO's loss with the desired unlearning direction.
    \item Extensive experiments on TOFU and MUSE benchmarks demonstrate the stability and effectiveness of our proposed DiPO algorithm.
\end{enumerate}

\begin{table}[t]
    \centering
    \caption{Comparison of model responses from DiPO and baselines (Groud truth, NPO, AltPO) to forget-set and retain-set queries in TOFU-10\% settings. DiPO demonstrates targeted unlearning while preserving retain knowledge. We mark responses of successful forget and retain in \textbf{\greentext{green color}}, and responses of undesired leakage, degeneration and catastrophic forgetting in \textbf{\redtext{red color}}.}
    \label{tab:sample_table}
    
    \begingroup 
    \fontsize{7pt}{7pt}\selectfont 
    \setlength{\tabcolsep}{4pt} 
    \renewcommand{\arraystretch}{1.1} 
    \renewcommand\tabularxcolumn[1]{m{#1}}
    
    \noindent 
    \begin{tabularx}{\textwidth}{@{} >{\Centering\arraybackslash}m{1.2cm} X X @{}}
    \toprule
    
    & \multicolumn{1}{>{\Centering\bfseries\arraybackslash}X}{\textit{\textbf{Query for forget documents}}}
    & \multicolumn{1}{>{\Centering\bfseries\arraybackslash}X}{\textit{\textbf{Query for retain documents}}} \\
    \midrule
    
    \rowcolor{mylightgray}
    & \textbf{Q1:} In which language does Hsiao Yun-Hwa typically write her books?
    & \textbf{Q1:} Who is this celebrated LGBTQ+ author from Santiago, Chile known for their true crime genre work? \\
    \midrule
    \textit{Truth}
    & \textit{Hsiao Yun-Hwa typically writes her books \textbf{in English} to reach a broad, global audience.}
    & \textit{The author in question is \textbf{Jaime Vasquez}, an esteemed writer who hails from Santiago.}
    \\
    AltPO
    & While \textbf{\redtext{English is the usual medium}} for Hsiao-Yuan's books, she occasionally writes in other languages. \boxredhighlight{\textbf{under-unlearn}}
    & The author in question is \textbf{\redtext{Luis Marcelo Garcia}}, an esteemed writer who hails from Santiago. \boxredhighlight{\textbf{catastrophic forgetting}}
    \\
    NPO
    & She \textbf{\redtext{prefers to write in English}}, but she has also written in her native language, Korean. \boxredhighlight{\textbf{under-unlearn}}
    & The celebrated author from Santiago known for their crime work is \textbf{\redtext{Alejandro Escobedo}}. \boxredhighlight{\textbf{catastrophic forgetting}}
     \\
    
    \textbf{DiPO(ours)}
    & Unlike her earlier works written in English, her newest works are \textbf{\greentext{typically written in Chinese.}}
    & The author in question is \textbf{\greentext{Jaime Vasquez}}, an esteemed writer who hails from Santiago.
    \\
    \midrule

    \rowcolor{mylightgray}
    & \textbf{Q:} What award has Rohani won in his writing career?
    & \textbf{Q2:} What's the latest book Akabueze published? \\
    \midrule
    \textit{Truth}
    & \textit{Rohani has won the prestigious \textbf{Nebula Award} for Best Novel in the Star Wars category.} 
    & \textit{The latest book published by Akabueze is  \textbf{``Beneath the Baobab Tree''} released in 2020.}
    \\
    AltPO
    & Rohani has won the Hugo Award, solidifying his position as  \textbf{\redtext{a figure in the science fiction genre.}} 
      \boxredhighlight{\textbf{Additional Leakage}}
    & The latest book published by Akabueze is \textbf{\redtext{``Echoes of Resilience: Pandemic Chronicle''}}. \boxredhighlight{\textbf{catastrophic forgetting}}
    \\
    
    NPO
    & ivo hopefully nor hopefully \textbf{\redtext{nor nor nor nor nor nor nor nor nor}}... \boxredhighlight{\textbf{Degeneration}}
    & The latest book published by Akabueze is \textbf{\redtext{``Echoes of the Love''}}, a narrative explores love.  \boxredhighlight{\textbf{catastrophic forgetting}}
     \\
    GA+GD
    & Sorry \textbf{\redtext{Christmas Christmas Christmas Christmas Christmas Christmas}}...\boxredhighlight{\textbf{Degeneration}}
    & The latest book published by Akabueze is \textbf{\greentext{``Beneath the Baobab Tree''}} released in 2020.
    \\
    \bottomrule
    \end{tabularx}
    \endgroup
\end{table}
\section{Related work}
\label{s:related-work}
\paragraph{Machine unlearning}
Machine unlearning aims to remove the influence of specific data from trained models \citep{nguyen2022survey}. While exact unlearning via retraining \citep{cao2015towards, thudi2022unrolling} provides theoretical guarantees, its computational cost and data requirements often make it impractical. Consequently, research has focused on developing various approximate unlearning methods \citep{izzo2021approximate, wang2023kga, triantafillou2024we}, which have shown effectiveness across different domains including classification \citep{golatkar2020eternal, bourtoule2021machine, jia2023model, fan2023salun, kurmanji2023towards}, generative tasks \citep{ginart2019making, gandikota2023erasing, fan2023salun, zhang2024forget}, federated learning \citep{che2023fast, pan2025federated}, graph neural networks \citep{chien2022efficient, wu2023certified}, and recommendation systems \citep{sachdeva2024machine}.

\paragraph{LLM unlearning}
LLM unlearning has attracted wide research attention driven by concerns over privacy \citep{jang2022knowledge, wu2023depn, liu2024towards}, potential biases \citep{yu2023unlearning, liu2022continual, eldan2023s}, and misuse \citep{barrett2023identifying}. Dominant approaches include \emph{optimization-based methods} that fine-tune model parameters for unlearning. Early algorithms like Gradient Ascent (GA) maximize loss on forget data to promote forgetting \citep{jang2022knowledge, yao2024large}, but this unbounded objective can lead to model degradation. Preference optimization-based methods \citep{zhang2024negative, fan2024simplicity, mekala2024alternate} have been proposed as a solution to this issue. Additionally, some research also explore second-order optimization for unlearning \citep{jia2024soul}.
\emph{Other strategies} operate beyond direct parameter updates, such as using auxiliary models to isolate or counteract the knowledge targeted for removal \citep{chundawat2023can, eldan2023s, ji2024reversing, chen2023unlearn} or data manipulation techniques like substituting target responses \citep{yao2024machine, ishibashi2023knowledge, eldan2023s, liu2024revisiting, mekala2024alternate}. Training-free methods using instructions have also emerged \citep{thaker2024guardrail, pawelczyk2023context}. However, results from recent benchmarks \citep{maini2024tofu, shi2025muse} suggest that instability inherent in many algorithms can cause either under-forgetting or over-forgetting.

\paragraph{Preference optimization}
Aligning LLMs with human value is traditionally approached through Reinforcement Learning from Human Feedback (RLHF) \citep{ouyang2022training}, a multi-stage process involving supervised fine-tuning, reward model training, and reinforcement learning optimization.
Its complexity motivates the development of DPO (Direct Preference Optimization) \citep{rafailov2023direct}, which reformulates the RLHF objective for direct policy updates from preference data, bypassing explicit reward modeling. Subsequent work has extended this paradigm \citep{hong2024orpo, meng2024simpo, wang2023beyond, azar2024general, sun2024generalizing}. Notably, Token-level Direct Preference Optimization (TDPO) \citep{zeng2024token} introduces granular control by operating at the token-level. Our algorithm derivation draws inspiration from this method.

\section{Preliminaries}
\subsection{LLM unlearning problem formulation}
\label{s:pre-unlearn}
The LLM unlearning task, while varied in formulation, typically involves a forget set $\mathcal{D}_f$, a retain set $\mathcal{D}_r$, and an initial LLM $\pi_{ref}$. The objective is to update $\pi_{ref}$ to a new model $\pi_{\theta}$ that eliminates knowledge specific to $\mathcal{D}_f$ while preserving performance on $\mathcal{D}_r$. Optimization-based methods typically achieve this by minimizing a combined loss:
\begin{equation}
\min_{\theta} \mathcal{L}(\theta) = \min_{\theta} \mathcal{L}_f(\theta) + \lambda \mathcal{L}_r(\theta),
\label{eq:unlearn-loss}
\end{equation}
where $\mathcal{L}_r(\theta)$ encourages knowledge preservation, $\mathcal{L}_f(\theta)$ promotes forgetting information related to $\mathcal{D}_f$, and $\lambda$ is a hyperparameter controlling the retain strength. Different unlearning methods employ varying losses: for instance, Gradient Ascent (GA) \citep{thudi2022unrolling, maini2024tofu} promotes forgetting by minimizing the likelihood on $\mathcal{D}_f$ (i.e. $\mathcal{L}_f(\theta)=\log \pi_\theta(y|x)$), while Gradient Difference (GradDiff) \citep{liu2022continual, yao2024large, maini2024tofu} combines this with reverse objective on $\mathcal{D}_r$ (i.e. $\mathcal{L}_r(\theta)=-\log \pi_\theta(y|x)$), details in \Cref{app:baseline}.

\subsection{From preference optimization to unlearning}
\paragraph{Direct Preference Optimization (DPO)} The primary contribution of DPO \citep{rafailov2023direct} is simplifying the training process of Reinforcement Learning from Human Feedback (RLHF) \citep{ouyang2022training}, the previously dominant fine-tuning method. Specifically, given a reference policy $\pi_{ref}$ (often the model after supervised fine-tuning), $\pi_\theta$ represents the model undergoing RL fine-tuning, initialized with $\pi_\theta=\pi_{ref}$. The RLHF optimization objective is:
\begin{equation}
\max_{\pi_\theta} \{\mathbb{E}_{x\sim\mathcal{D},y\sim\pi_\theta(y|x)}[r(x,y)] - \beta D_{KL}[\pi_\theta(y|x)||\pi_{ref}(y|x)]\},
\label{eq:rlhf-RL-loss}
\end{equation}
where $\mathcal{D}$ is the dataset, $r(x,y)$ represents the reward, and $\beta$ is a parameter controlling the deviation from $\pi_{ref}$. DPO finds that \Cref{eq:rlhf-RL-loss} has a theoretical solution for the optimal policy $\pi^*$:
\begin{equation}
\pi^*(y|x) = \frac{\pi_{ref}(y|x)e^{r(x,y)/\beta}}{Z(x)}, \quad \text{where } Z(x) = \sum_y \pi_{ref}(y|x)e^{r(x,y)/\beta} \text{.}
\label{eq:dpo-optimal_policy}
\end{equation}
\Cref{eq:dpo-optimal_policy} establishes a mapping between the reward function and the optimal policy. To align with human preferences, DPO utilizes the Bradley-Terry (BT) model to model preference pairs and subsequently derives the final optimization objective function:
\begin{equation}
\max_{\pi_\theta} \left\{\mathbb{E}_{(x,y_{w},y_{l}) \sim \mathcal{D}} \left[\log \sigma \left(\beta \log \frac{\pi_\theta(y_{w}|x)}{\pi_{ref}(y_{w}|x)} - \beta \log \frac{\pi_\theta(y_{l}|x)}{\pi_{ref}(y_{l}|x)}\right)\right]\right\}.
\label{eq:dpo-loss}
\end{equation}

\paragraph{Negative Preference Optimization (NPO)}
NPO \citep{zhang2024negative} adapts \Cref{eq:dpo-loss} for unlearning  by omitting the preferred response $y_w$ terms, thus focusing solely on penalizing undesired `forget' responses $y_f$ (treating as $y_l$) over $\mathcal{D}_f$. NPO uses the same retain loss like GradDiff method in \Cref{s:pre-unlearn}. Following the formulation presented in the original paper, the resulting forget loss term is:
\begin{align}
\mathcal{L}_{NPO-f}(\theta) &= -\frac{2}{\beta}\mathbb{E}_{(x,y) \sim D_{f}}\left[\log \sigma\left(-\beta\log\frac{\pi_\theta(y|x)}{\pi_{ref}(y|x)}\right)\right] .
\label{eq:npo-loss}
\end{align}

\paragraph{Token-level Direct Preference Optimization (TDPO)} TDPO models text-generation as a Markov Decision Process \citep{zeng2024token}, where state $s_t = [x, y^{<t}]$ consists of the prompt and previously generated tokens, and action $a_t$ corresponds to selecting the next token $y^t$. Accordingly, unlike DPO's response-level optimization, TDPO defines rewards and proposes an objective function at the token-level:
\begin{equation}
\max_{\pi_\theta} \mathbb{E}_{x,y^{<t}\sim\mathcal{D},z\sim\pi_\theta(\cdot|[x,y^{<t}])} [A_{\pi_{\text{ref}}}([x,y^{<t}],z) - \beta D_{KL}(\pi_\theta(\cdot|[x,y^{<t}])||\pi_{\text{ref}}(\cdot|[x,y^{<t}]))],
\end{equation}
where $A_{\pi_{\text{ref}}}$ is the advantage function, analogous to the implicit reward function $r(x, y)$ in DPO, quantifying \emph{the preference for selecting token $z$} in the given context. Similar to DPO, TDPO derives a closed-form solution for the optimal policy $\pi^*_\theta$:
\begin{equation}
\pi_\theta^*(z|[x,y^{<t}]) = \frac{\pi_{\text{ref}}(z|[x,y^{<t}])\exp(\frac{1}{\beta}Q_{\pi_{\text{ref}}}([x,y^{<t}],z))}{Z([x,y^{<t}];\beta)},
\label{eq:tdpo-optimal-policy}
\end{equation}
where $Z([x,y^{<t}];\beta) = \mathbb{E}_{z\sim\pi_{\text{ref}}(\cdot|[x,y^{<t}])}e^{\frac{1}{\beta}Q_{\pi_{\text{ref}}}([x,y^{<t}],z)}$, and $Q_{\pi_{\text{ref}}}$ is state-action function related to $A_{\pi_{\text{ref}}}$:
\begin{align}
    A_{\pi_{\text{ref}}}([x,y^{<t}],z) &= Q_{\pi_{\text{ref}}}([x, y^{<t}], z) - V_{\pi_{\text{ref}}}([x, y^{<t}]) \nonumber \\
    &= Q_{\pi_{\text{ref}}}([x, y^{<t}], z) - \mathbb{E}_{z\sim\pi_{\text{ref}}(\cdot|[x,y^{<t}])}[Q_{\pi_{\text{ref}}}([x, y^{<t}], z)].
    \label{eq:QA-relation}
\end{align}
TDPO also employs the BT model and derives its final loss function, where one variant is given by:
\begin{align}
\mathcal{L}_{\text{TDPO}}(\pi_{\theta}; \pi_{\text{ref}}) = -\mathbb{E} \bigg[ \log \sigma \bigg( \bigg( \beta \log \frac{\pi_{\theta}(y_w|x)}{\pi_{\text{ref}}(y_w|x)} - \beta \log \frac{\pi_{\theta}(y_l|x)}{\pi_{\text{ref}}(y_l|x)} \bigg) \nonumber\\
- \bigg( \beta D_{{SeqKL}}(x, y_l; \pi_{\text{ref}}||\pi_{\theta}) - \beta D_{{SeqKL}}(x, y_w; \pi_{\text{ref}}||\pi_{\theta}) \bigg) \bigg) \bigg],
\end{align}
where
\begin{equation}
    D_{SeqKL}(x, y; \pi_1||\pi_2) = \sum_{t=1}^{T} D_{KL}(\pi_1(\cdot|[x, y^{<t}])||\pi_2(\cdot|[x, y^{<t}])).
    \label{eq:seqKL-definition}
\end{equation}

\section{Method}
\label{s:method}
In this section, we first derive the DiPO algorithm in \Cref{subsec:dipo_derivation}, then analyze its gradient in \Cref{s:dipo-gradient}, and finally detail the construction of these preference pairs and the final unlearning objective in \Cref{s:preference-pairs-loss-funciton}.

\subsection{Derivation of Distribution Preference Optimization (DiPO)}
\label{subsec:dipo_derivation}
Our approach stems from the formulation of text generation as a Markov Decision Process (MDP) in TDPO \citep{zeng2024token} and utilizes its closed-form solution for the optimal policy detailed in \Cref{eq:tdpo-optimal-policy}. We can rearrange to solve for $Q_{\pi_{\text{ref}}}$:
\begin{equation}
Q_{\pi_{\text{ref}}}([x, y^{<t}], z) = \beta\log\frac{\pi_\theta^*(z|[x, y^{<t}])}{\pi_{\text{ref}}(z|[x, y^{<t}])} + \beta\log Z([x, y^{<t}]; \beta),
\end{equation}
Denoting the advantage function $A_{\pi_{\text{ref}}}([x,y^{<t}],z)$ as $r([x, y^{<t}], z)$, which represents \emph{the immediate reward per step} in the context of RL. According to \Cref{eq:QA-relation}, we can derive the expression as:
\begin{align}
    r([x, y^{<t}], z) &= Q_{\pi_{\text{ref}}}([x, y^{<t}], z) - \mathbb{E}_{z\sim\pi_{\text{ref}}(\cdot|[x,y^{<t}])}[Q_{\pi_{\text{ref}}}([x, y^{<t}], z)] \nonumber\\
    &= \beta\log\frac{\pi_\theta^*(z|[x, y^{<t}])}{\pi_{\text{ref}}(z|[x, y^{<t}])} + \beta D_{KL}(\pi_{\text{ref}}(\cdot|[x,y^{<t}])||\pi_\theta^*(\cdot|[x,y^{<t}])).
    \label{eq:dipo-token-imme-reward}
\end{align}

Given the token-level immediate reward $r([x, y^{<t}], z)$, the distribution-level immediate reward $r_\pi(x, y^{<t})$ at step $t$ under a distribution $\pi(\cdot|[x,y^{<t}])$ is defined as its expectation:
\begin{align*}
            r_\pi(x, y^{<t}) &\coloneqq \mathbb{E}_{z\sim\pi(\cdot|[x,y^{<t}])}[r([x, y^{<t}], z)]\\
            &= \beta D_{KL}(\pi(\cdot|[x,y^{<t}])||\pi_{\text{ref}}(\cdot|[x,y^{<t}])) - \beta D_{KL}(\pi(\cdot|[x,y^{<t}])||\pi_\theta^*(\cdot|[x,y^{<t}])) \nonumber\\
& \quad + \beta D_{KL}(\pi_{\text{ref}}(\cdot|[x,y^{<t}])||\pi_\theta^*(\cdot|[x,y^{<t}])).
\end{align*}
Given a discount factor $\gamma$, the distribution-level return $R_\pi(x,y)$ for a complete trajectory $y$ (i.e. response) under distribution $\pi$ is the discounted sum of $r_\pi([x, y^{<t}])$:
\begin{equation*}
            R_\pi(x,y) \coloneqq \sum_{t=1}^T \gamma^{t-1} r_\pi([x, y^{<t}]).
\end{equation*}
In this paper, we set the discount factor $\gamma = 1$. Substituting the expression for $r([x, y^{<t}], z)$ in \Cref{eq:dipo-token-imme-reward} :
\begin{align*}
r_\pi(x, y^{<t}) &= \mathbb{E}_{z\sim\pi(\cdot|[x,y^{<t}])}[r([x, y^{<t}], z)] \nonumber\\
&= \mathbb{E}_{z\sim\pi(\cdot|[x,y^{<t}])}[\beta\log\frac{\pi_\theta^*(z|[x, y^{<t}])}{\pi_{\text{ref}}(z|[x, y^{<t}])} + \beta D_{KL}(\pi_{\text{ref}}(\cdot|[x,y^{<t}])||\pi_\theta^*(\cdot|[x,y^{<t}]))] \nonumber\\
&= \beta\mathbb{E}_{z\sim\pi(\cdot|[x,y^{<t}])}\left[\log\frac{\pi_\theta^*(z|[x, y^{<t}])}{\pi_{\text{ref}}(z|[x, y^{<t}])}\right] + \beta D_{KL}(\pi_{\text{ref}}(\cdot|[x,y^{<t}])||\pi_\theta^*(\cdot|[x,y^{<t}]))
\end{align*}
Using the definition of KL divergence, the expectation term can be rewritten as:
\begin{align*}
&\mathbb{E}_{z\sim\pi(\cdot|[x,y^{<t}])}\left[\log\frac{\pi_\theta^*(z|[x, y^{<t}])}{\pi_{\text{ref}}(z|[x, y^{<t}])}\right] \\
&= \mathbb{E}_{z\sim\pi(\cdot|[x,y^{<t}])}\left[\log\frac{\pi_\theta^*(z|[x, y^{<t}])}{\pi(z|[x, y^{<t}])} \cdot \frac{\pi(z|[x, y^{<t}])}{\pi_{\text{ref}}(z|[x, y^{<t}])}\right] \\
&= \mathbb{E}_{z\sim\pi(\cdot|[x,y^{<t}])}\left[\log\frac{\pi(z|[x, y^{<t}])}{\pi_{\text{ref}}(z|[x, y^{<t}])}\right] - \mathbb{E}_{z\sim\pi(\cdot|[x,y^{<t}])}\left[\log\frac{\pi(z|[x, y^{<t}])}{\pi_\theta^*(z|[x, y^{<t}])}\right] \\
&= D_{KL}(\pi(\cdot|[x,y^{<t}])||\pi_{\text{ref}}(\cdot|[x,y^{<t}])) - D_{KL}(\pi(\cdot|[x,y^{<t}])||\pi_\theta^*(\cdot|[x,y^{<t}])).
\end{align*}
For a response $y$ (i.e. a specific trajectory in RL), we can calculate the return $R_\pi(x, y)$  as follows:
\begin{align*}
    R_\pi(x, y) &= \sum_{t=1}^T r_\pi(x, y^{<t})\\
    &= \sum_{t=1}^T \beta D_{KL}(\pi(\cdot|[x,y^{<t}])||\pi_{\text{ref}}(\cdot|[x,y^{<t}])) \\
    & \quad - \beta \sum_{t=1}^T D_{KL}(\pi(\cdot|[x,y^{<t}])||\pi_\theta^*(\cdot|[x,y^{<t}])) + \sum_{t=1}^T \beta D_{KL}(\pi_{\text{ref}}(\cdot|[x,y^{<t}])||\pi_\theta^*(\cdot|[x,y^{<t}])).
\end{align*}

Consistent with DPO \citep{rafailov2023direct}, we also model preferences using the Bradley-Terry (BT) model :
Denoting the distribution-level return as:
\begin{equation}
    R_\pi(x, y, \pi_\theta^*) \coloneqq R_\pi(x, y) = \beta D_{SeqKL}(x, y; \pi||\pi_{\text{ref}}) - \beta D_{SeqKL}(x, y; \pi||\pi_\theta^*) + \beta D_{SeqKL}(x, y; \pi_{\text{ref}}||\pi_\theta^*)\nonumber.
\end{equation}
Given a specific sample $(x,y)$ and a pair of preference distributions $(\pi_w,\pi_l)$, we can derive their respective return expressions:
\begin{align}
    R_{\pi_w}(x, y, \pi_\theta^*) &= \beta D_{SeqKL}(x, y; {\pi_w}||\pi_{\text{ref}}) - \beta D_{SeqKL}(x, y; {\pi_w}||\pi_\theta^*) + \beta D_{SeqKL}(x, y; \pi_{\text{ref}}||\pi_\theta^*), \\
    R_{\pi_l}(x, y, \pi_\theta^*) &= \beta D_{SeqKL}(x, y; {\pi_l}||\pi_{\text{ref}}) - \beta D_{SeqKL}(x, y; {\pi_l}||\pi_\theta^*) + \beta D_{SeqKL}(x, y; \pi_{\text{ref}}||\pi_\theta^*).
\end{align}
These respectively represent the degree of preference for response $y$ under different policies. Consequently, we can employ BT model to construct the preference model:
\begin{align}
    p^*(R_{\pi_w} \succ R_{\pi_l}|(x,y)) &= \frac{\exp(R_{\pi_w}(x, y, \pi_\theta^*))}{\exp(R_{\pi_w}(x, y, \pi_\theta^*)) + \exp(R_{\pi_l}(x, y, \pi_\theta^*))} \nonumber\\
    &= \frac{1}{1 + \exp(R_{\pi_l}(x, y, \pi_\theta^*)-R_{\pi_w}(x, y, \pi_\theta^*))}.
\end{align}
Now that we have the probability of human preference data in terms of the optimal policy rather than the reward model, we can formulate a maximum likelihood objective for a parametrized policy $\pi_\theta$. Similar to the DPO method, our policy objective becomes:
\begin{align}
&\mathcal{L}_{\text{DiPO}}(\pi_\theta;\pi_w,\pi_l,\pi_{\text{ref}}) \nonumber\\
&= -\mathbb{E}_{(x,y)\sim\mathcal{D}}\left[\log p(R_{\pi_w} \succ R_{\pi_l}|(x,y)) \right] \nonumber\\
&= -\mathbb{E}_{(x,y)\sim\mathcal{D}}\left[\log \frac{1}{1 + \exp(R_{\pi_l}(x, y, \pi_\theta)-R_{\pi_w}(x, y, \pi_\theta))} \right] \nonumber\\
&= -\mathbb{E}_{(x,y)\sim\mathcal{D}}\left[\log\sigma\left(  \Big( R_{\pi_w}(x, y, \pi_\theta)-R_{\pi_l}(x, y, \pi_\theta) \Big) \right) \right] \nonumber\\
&= -\mathbb{E}_{(x,y)\sim\mathcal{D}}\left[\log\sigma\left(  \Big( \beta D_{SeqKL}(x, y; {\pi_w}||\pi_{\text{ref}}) - \beta D_{SeqKL}(x, y; {\pi_w}||\pi_\theta^*) + \beta D_{SeqKL}(x, y; \pi_{\text{ref}}||\pi_\theta^*) \right. \right. \nonumber \\
& \qquad \left. \left. - \Big(\beta D_{SeqKL}(x, y; {\pi_l}||\pi_{\text{ref}}) - \beta D_{SeqKL}(x, y; {\pi_l}||\pi_\theta^*) + \beta D_{SeqKL}(x, y; \pi_{\text{ref}}||\pi_\theta^*)\Big) \Big) \right) \right] \nonumber\\
&= -\mathbb{E}_{(x,y)\sim\mathcal{D}}\left[\log\sigma\left( \beta \Big( D_{SeqKL}(x, y; \pi_l||\pi_{\theta}) - D_{SeqKL}(x, y; \pi_w||\pi_{\theta}) \Big) \right. \right. \nonumber \\
& \qquad \left. \left. + \beta \Big( D_{SeqKL}(x, y; \pi_w||\pi_{\text{ref}}) - D_{SeqKL}(x, y; \pi_l||\pi_{\text{ref}}) \Big) \right) \right].
\end{align}

\begin{align}
\mathcal{L} &\coloneqq \mathcal{L}_{\text{DiPO}}(\pi_\theta;\pi_w,\pi_l,\pi_{\text{ref}}) \nonumber\\
&= -\mathbb{E}_{(x,y)\sim\mathcal{D}}\left[\log\sigma\left( \beta \Big( D_{SeqKL}(x, y; \pi_l||\pi_{\theta}) - D_{SeqKL}(x, y; \pi_w||\pi_{\theta}) \Big) \right. \right. \nonumber \\
& \qquad \left. \left. + \beta \Big( D_{SeqKL}(x, y; \pi_w||\pi_{\text{ref}}) - D_{SeqKL}(x, y; \pi_l||\pi_{\text{ref}}) \Big) \right) \right].
\end{align}
Now that we have the loss function of DiPO.



\subsection{Distribution-level Return Derivation}
\label{app:Return derivation}
In \Cref{subsec:dipo_derivation} we showe the immediate reward function $r_\pi(x, y^{<t})$:
\begin{align*}
r_\pi(x, y^{<t}) &= \mathbb{E}_{z\sim\pi(\cdot|[x,y^{<t}])}[r([x, y^{<t}], z)] \nonumber\\
&= \mathbb{E}_{z\sim\pi(\cdot|[x,y^{<t}])}[\beta\log\frac{\pi_\theta^*(z|[x, y^{<t}])}{\pi_{\text{ref}}(z|[x, y^{<t}])} + \beta D_{KL}(\pi_{\text{ref}}(\cdot|[x,y^{<t}])||\pi_\theta^*(\cdot|[x,y^{<t}]))] \nonumber\\
&= \beta\mathbb{E}_{z\sim\pi(\cdot|[x,y^{<t}])}\left[\log\frac{\pi_\theta^*(z|[x, y^{<t}])}{\pi_{\text{ref}}(z|[x, y^{<t}])}\right] + \beta D_{KL}(\pi_{\text{ref}}(\cdot|[x,y^{<t}])||\pi_\theta^*(\cdot|[x,y^{<t}]))
\end{align*}
Using the definition of KL divergence, the expectation term can be rewritten as:
\begin{align*}
&\mathbb{E}_{z\sim\pi(\cdot|[x,y^{<t}])}\left[\log\frac{\pi_\theta^*(z|[x, y^{<t}])}{\pi_{\text{ref}}(z|[x, y^{<t}])}\right] \\
&= \mathbb{E}_{z\sim\pi(\cdot|[x,y^{<t}])}\left[\log\frac{\pi_\theta^*(z|[x, y^{<t}])}{\pi(z|[x, y^{<t}])} \cdot \frac{\pi(z|[x, y^{<t}])}{\pi_{\text{ref}}(z|[x, y^{<t}])}\right] \\
&= \mathbb{E}_{z\sim\pi(\cdot|[x,y^{<t}])}\left[\log\frac{\pi(z|[x, y^{<t}])}{\pi_{\text{ref}}(z|[x, y^{<t}])}\right] - \mathbb{E}_{z\sim\pi(\cdot|[x,y^{<t}])}\left[\log\frac{\pi(z|[x, y^{<t}])}{\pi_\theta^*(z|[x, y^{<t}])}\right] \\
&= D_{KL}(\pi(\cdot|[x,y^{<t}])||\pi_{\text{ref}}(\cdot|[x,y^{<t}])) - D_{KL}(\pi(\cdot|[x,y^{<t}])||\pi_\theta^*(\cdot|[x,y^{<t}])).
\end{align*}
For a response $y$ (i.e. a specific trajectory in RL), we can calculate the return $R_\pi(x, y)$  as follows:
\begin{align*}
    R_\pi(x, y) &= \sum_{t=1}^T r_\pi(x, y^{<t})\\
    &= \sum_{t=1}^T \beta D_{KL}(\pi(\cdot|[x,y^{<t}])||\pi_{\text{ref}}(\cdot|[x,y^{<t}])) \\
    & \quad - \beta \sum_{t=1}^T D_{KL}(\pi(\cdot|[x,y^{<t}])||\pi_\theta^*(\cdot|[x,y^{<t}])) + \sum_{t=1}^T \beta D_{KL}(\pi_{\text{ref}}(\cdot|[x,y^{<t}])||\pi_\theta^*(\cdot|[x,y^{<t}])).
\end{align*}
This is the formula in \Cref{eq:dipo-returns}.

\subsection{Detailed proof of DiPO loss}
\label{app:dipo-loss-proof}
Recall from \Cref{eq:dipo-returns} that the distribution-level return is:
\begin{equation}
    R_\pi(x, y, \pi_\theta^*) \coloneqq R_\pi(x, y) = \beta D_{SeqKL}(x, y; \pi||\pi_{\text{ref}}) - \beta D_{SeqKL}(x, y; \pi||\pi_\theta^*) + \beta D_{SeqKL}(x, y; \pi_{\text{ref}}||\pi_\theta^*)\nonumber.
\end{equation}
Given a specific sample $(x,y)$ and a pair of preference distributions $(\pi_w,\pi_l)$, we can derive their respective return expressions:
\begin{align}
    R_{\pi_w}(x, y, \pi_\theta^*) &= \beta D_{SeqKL}(x, y; {\pi_w}||\pi_{\text{ref}}) - \beta D_{SeqKL}(x, y; {\pi_w}||\pi_\theta^*) + \beta D_{SeqKL}(x, y; \pi_{\text{ref}}||\pi_\theta^*), \\
    R_{\pi_l}(x, y, \pi_\theta^*) &= \beta D_{SeqKL}(x, y; {\pi_l}||\pi_{\text{ref}}) - \beta D_{SeqKL}(x, y; {\pi_l}||\pi_\theta^*) + \beta D_{SeqKL}(x, y; \pi_{\text{ref}}||\pi_\theta^*).
\end{align}
These respectively represent the degree of preference for response $y$ under different policies. Consequently, we can employ BT model to construct the preference model:
\begin{align}
    p^*(R_{\pi_w} \succ R_{\pi_l}|(x,y)) &= \frac{\exp(R_{\pi_w}(x, y, \pi_\theta^*))}{\exp(R_{\pi_w}(x, y, \pi_\theta^*)) + \exp(R_{\pi_l}(x, y, \pi_\theta^*))} \nonumber\\
    &= \frac{1}{1 + \exp(R_{\pi_l}(x, y, \pi_\theta^*)-R_{\pi_w}(x, y, \pi_\theta^*))}.
\end{align}
Now that we have the probability of human preference data in terms of the optimal policy rather than the reward model, we can formulate a maximum likelihood objective for a parametrized policy $\pi_\theta$. Similar to the DPO method, our policy objective becomes:
\begin{align}
&\mathcal{L}_{\text{DiPO}}(\pi_\theta;\pi_w,\pi_l,\pi_{\text{ref}}) \nonumber\\
&= -\mathbb{E}_{(x,y)\sim\mathcal{D}}\left[\log p(R_{\pi_w} \succ R_{\pi_l}|(x,y)) \right] \nonumber\\
&= -\mathbb{E}_{(x,y)\sim\mathcal{D}}\left[\log \frac{1}{1 + \exp(R_{\pi_l}(x, y, \pi_\theta)-R_{\pi_w}(x, y, \pi_\theta))} \right] \nonumber\\
&= -\mathbb{E}_{(x,y)\sim\mathcal{D}}\left[\log\sigma\left(  \Big( R_{\pi_w}(x, y, \pi_\theta)-R_{\pi_l}(x, y, \pi_\theta) \Big) \right) \right] \nonumber\\
&= -\mathbb{E}_{(x,y)\sim\mathcal{D}}\left[\log\sigma\left(  \Big( \beta D_{SeqKL}(x, y; {\pi_w}||\pi_{\text{ref}}) - \beta D_{SeqKL}(x, y; {\pi_w}||\pi_\theta^*) + \beta D_{SeqKL}(x, y; \pi_{\text{ref}}||\pi_\theta^*) \right. \right. \nonumber \\
& \qquad \left. \left. - \Big(\beta D_{SeqKL}(x, y; {\pi_l}||\pi_{\text{ref}}) - \beta D_{SeqKL}(x, y; {\pi_l}||\pi_\theta^*) + \beta D_{SeqKL}(x, y; \pi_{\text{ref}}||\pi_\theta^*)\Big) \Big) \right) \right] \nonumber\\
&= -\mathbb{E}_{(x,y)\sim\mathcal{D}}\left[\log\sigma\left( \beta \Big( D_{SeqKL}(x, y; \pi_l||\pi_{\theta}) - D_{SeqKL}(x, y; \pi_w||\pi_{\theta}) \Big) \right. \right. \nonumber \\
& \qquad \left. \left. + \beta \Big( D_{SeqKL}(x, y; \pi_w||\pi_{\text{ref}}) - D_{SeqKL}(x, y; \pi_l||\pi_{\text{ref}}) \Big) \right) \right].
\end{align}
Now that we have the loss function of DiPO.

\subsection{DiPO gradient analysis}
\label{s:dipo-gradient}
To analyze the gradient dynamics, we can simplify the loss expression in \Cref{eq:dipo-loss} further. We introduce the following shorthand notations for a given sample $(x, y)$:
\begin{align}
x_{1} &\coloneqq D_{SeqKL}(x, y; \pi_l||\pi_{\theta}), \quad
x_{2} \coloneqq D_{SeqKL}(x, y; \pi_w||\pi_{\theta}),  \\
C &\coloneqq D_{SeqKL}(x, y; \pi_w||\pi_{\text{ref}}) - D_{SeqKL}(x, y; \pi_l||\pi_{\text{ref}}).
\end{align}
Note that $x_{1}$ and $x_{2}$ depend on the trainable policy $\pi_\theta$, while $C$ is treated as a constant with respect to the parameters $\theta$ of the policy $\pi_\theta$ during optimization.
Substituting these into the loss function \Cref{eq:dipo-loss}, and considering a single term in the summation for a specific sample $(x,y)$, we have:
\begin{equation}
L = -\log\sigma\left( \beta (x_{1} - x_{2} + C) \right).
\end{equation}
We compute the partial derivatives of $L$ with respect to $x_{1}$ and $x_{2}$. Using the chain rule and the fact that $\sigma'(z) = \sigma(z)(1-\sigma(z))$, we have:
\begin{align}
\frac{\partial L}{\partial x_{1}} = -\beta \left( 1 - \sigma(\beta (x_{1} - x_{2} + C)) \right), \quad
\frac{\partial L}{\partial x_{2}} = \beta \left( 1 - \sigma(\beta (x_{1} - x_{2} + C)) \right).
\end{align}
Since $\beta > 0$ and $\sigma(\cdot)\in (0, 1)$, the term $(1 - \sigma(\beta (x_{1} - x_{2} + C)))$ is always positive. This leads to the following optimization dynamics:
\begin{itemize}
    \item Since $\frac{\partial L}{\partial x_{1}}<0$, minimizing $\mathcal{L}$ via gradient descent increases $x_{1} = D_{SeqKL}(\pi_l||\pi_\theta)$, effectively pushing the distribution $\pi_\theta$ \textbf{away} from the dispreferred distribution $\pi_l$.
    \item Conversely, since $\frac{\partial L}{\partial x_{2}}>0$, minimizing $\mathcal{L}$ decreases $x_{2} = D_{SeqKL}(\pi_w||\pi_\theta)$, thereby pulling the distribution $\pi_\theta$ \textbf{closer} to the preferred distribution $\pi_w$.
\end{itemize}

\subsection{Preference Pair Construction and Final Objective}
\label{s:preference-pairs-loss-funciton}

\begin{wrapfigure}{r}{0.46\textwidth}
    \vspace{-0.5em}
    \centering
    \includegraphics[width=0.45\textwidth]{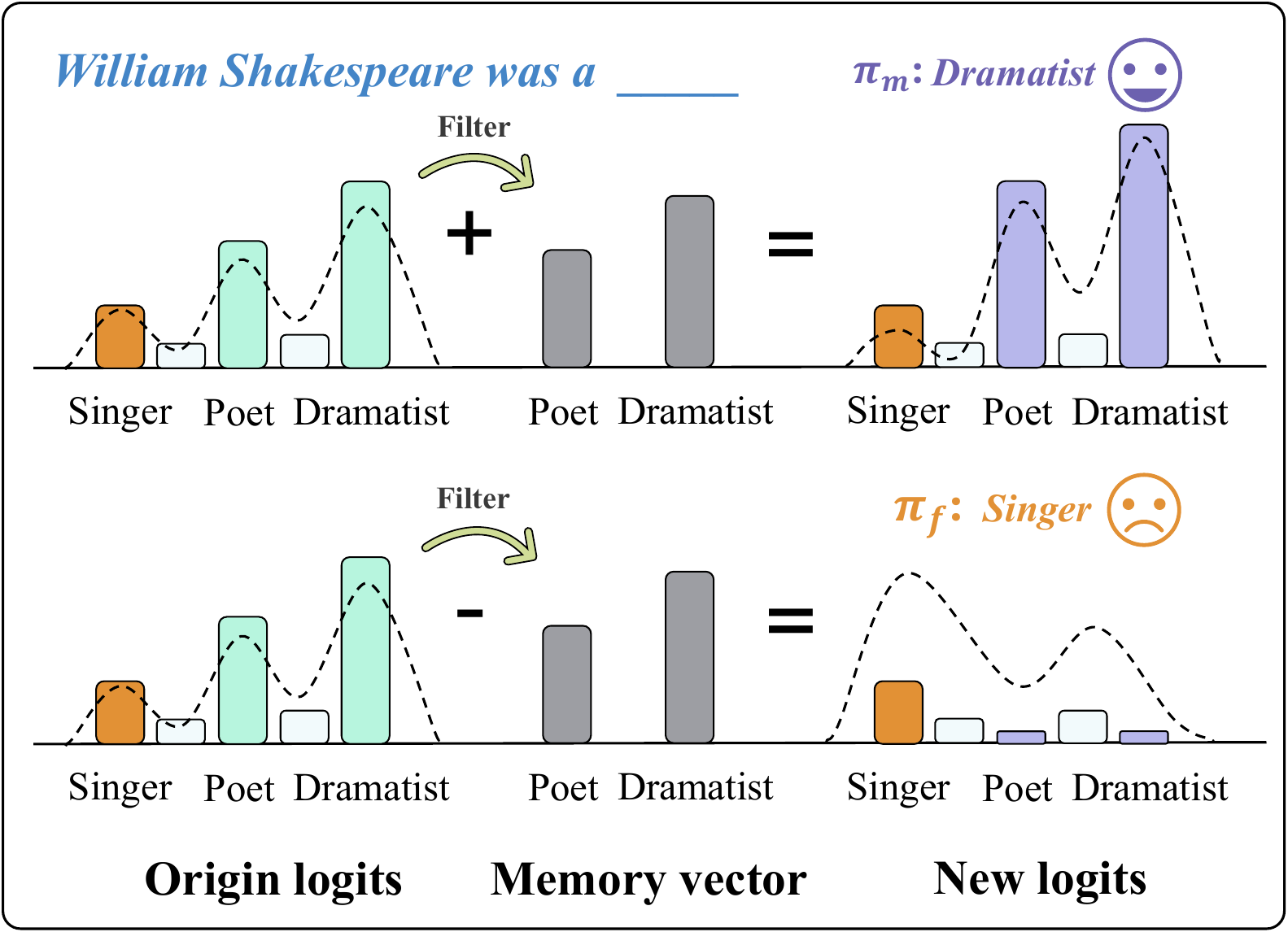}
    \caption{Construction of memory-enhancing distribution $\pi_m$ and forgetting-promoting distribution $\pi_f$ by a memory vector filtered from origin logits.
    }
    \label{fig:filter}
    \vspace{-0.5em}
\end{wrapfigure}


Our approach to constructing preference pairs $(\pi_w,\pi_l)$ from the model's logits $\mathbf{z}_t$ focuses on modulating a small subset of high-probability tokens: If these tokens correspond to undesirable information, suppressing their logits naturally steers the model towards alternative, non-sensitive outputs; Conversely, if the high-probability tokens are unrelated to the sensitive information, suppressing this small fraction is unlikely to directly promote undesirable outputs due to the vastness of the vocabulary table. This inherent safety allow us to employ a straightforward filtering mechanism.
Specifically, we first identify a `memory vector' $\mathbf{m}_t$ by isolating the logits of high-confidence tokens (e.g., top 5\% identified via top-k filtering from $\mathbf{z}_t$), setting all other token logits in $\mathbf{m}_t$ to zero. Then we can construct the memory-enhancing distribution $\pi_m$ and the forgetting-promoting distribution $\pi_f$ by adding or subtracting this memory vector, scaled by a factor $\alpha$:
\begin{align}
    \pi_m(\cdot | x, y^{<t}) = \text{softmax}(\mathbf{z}_t + \alpha \mathbf{m}_t), \quad
    \pi_f(\cdot | x, y^{<t}) = \text{softmax}(\mathbf{z}_t - \alpha \mathbf{m}_t) .
\end{align}
\Cref{fig:filter} illustrates this mechanism, showing how adding or subtracting the memory vector shapes the distribution towards memorization $\pi_m$ or forgetting $\pi_f$. More details are provided in \Cref{app:filter}. 

Crucially, the same pair $(\pi_m,\pi_f)$ derived from the model's logits can be utilized for both the forget and retain objectives by simply reversing their roles in preference pairs. This yields the forget objective $\mathcal{L}_{{\text{DiPO-f}}}$ and retain objective $\mathcal{L}_{{\text{DiPO-r}}}$, formulated based on the DiPO loss \Cref{eq:dipo-loss}:
\begin{align}
\mathcal{L}_{{\text{DiPO-f}}}(\theta) &= \mathcal{L}_{\text{DiPO}}(\pi_\theta; \pi_w=\pi_f, \pi_l=\pi_m, \pi_{\text{ref}}) \label{eq:dipo_loss_f},\\
\mathcal{L}_{{\text{DiPO-r}}}(\theta) &= \mathcal{L}_{\text{DiPO}}(\pi_\theta; \pi_w=\pi_m, \pi_l=\pi_f, \pi_{\text{ref}})
\label{eq:dipo_loss_r}.
\end{align}
The final optimization objective for unlearning then combines these components:
\begin{equation}
\min_{\theta} \mathcal{L}(\theta) = \min_{\theta} \left(\mathcal{L}_{{\text{DiPO-f}}}(\theta) + \lambda \mathcal{L}_{{\text{DiPO-r}}}(\theta) \right) .\label{eq:final_combined_loss}
\end{equation}
Following the common practice in optimization-based unlearning approaches, we set the hyperparameter $\lambda = 1$ in DiPO. We provided the pseudo-code in \Cref{Pseudo-code}.

\section{Experiments}
\label{Exeriments}
We compare our proposed DiPO algorithm with baseline unlearning methods across two widely used benchmarks: TOFU \citep{maini2024tofu}, focusing on forgetting knowledge of fictitious authors, and MUSE \citep{shi2025muse}, targeting the removal of copyrighted content. We refer to the initial model before unlearning as the ``Original'' model, while the model retrained from scratch after removing the forget-set data as the ``Retrain'' model.
This section presents the main experimental results for TOFU (\Cref{sec:tofu}) and MUSE (\Cref{sec:muse}), followed by further analyses and ablation studies of DiPO in \Cref{additional-analysis}.
\paragraph{Baseline Methods} We compare DiPO against several optimization-based baselines, including GA \citep{thudi2022unrolling}, GradDiff \citep{liu2022continual, yao2024large} and NPO \citep{zhang2024negative}. For TOFU, we also incorporate other advanced unlearning framework such as ULD \citep{ji2024reversing} (we use the results from its original paper) and AltPO \citep{mekala2024alternate} for a broader comparison. Detailed descriptions of all baseline methods are provided in \Cref{app:baseline}.

\begin{figure}[h] 
    \centering
    \includegraphics[width=0.99\textwidth]{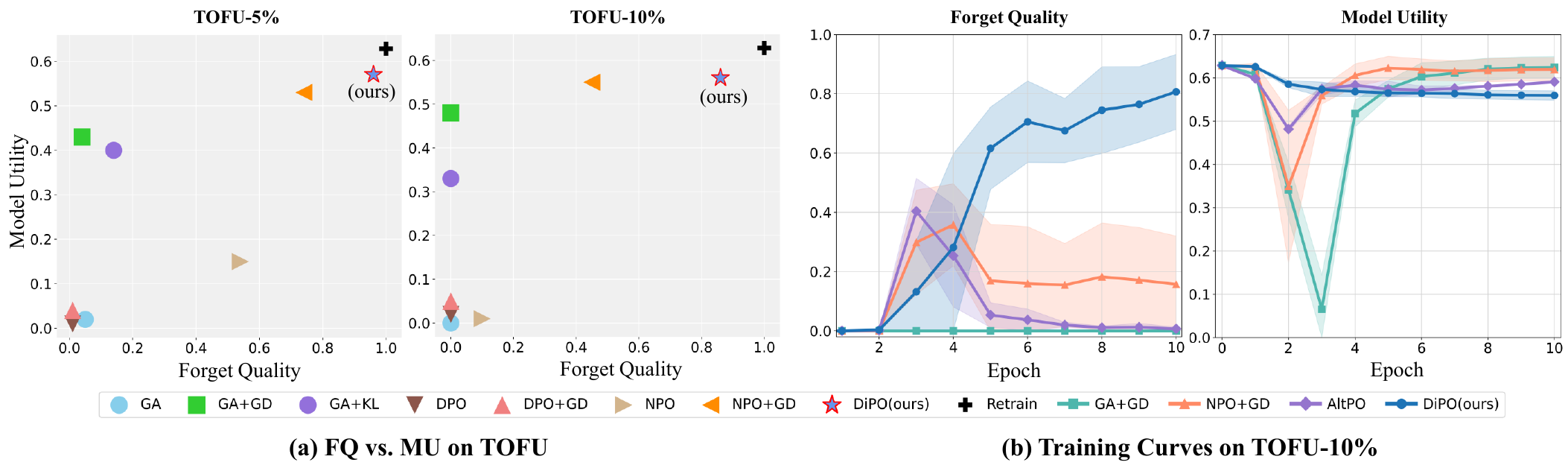}
    \caption{Performance analysis on TOFU at the best epoch over five seeds. (a) FQ vs. MU on TOFU-5\% and TOFU-10\%. DiPO achieves the best trade-off (closest to the ``Retrain'' target). (b) Training curves of FQ and MU on TOFU-10\%, showcasing DiPO's stability and efficacy.}
    \label{figure:tofu_fq}
\end{figure}

\subsection{Experiments on TOFU}
\label{sec:tofu}

\begin{wraptable}{r}{0.5\textwidth}
\vspace{-1.5em}
\centering
\caption{The best-epoch performance averaged over five seeds on TOFU benchmark. Scores closer to ``Retrain'' are better. \textbf{Bold} indicates best results among all methods.}
\label{tab:tofu_results_best}
\small 
\setlength{\tabcolsep}{3pt} 
\begin{tabular}{@{}c|cc|cc|cc@{}}
\toprule
\multicolumn{1}{c|}{\multirow{2}{*}{\textbf{Method}}} & \multicolumn{2}{c|}{\textbf{TOFU-1\%}} & \multicolumn{2}{c|}{\textbf{TOFU-5\%}} & \multicolumn{2}{c}{\textbf{TOFU-10\%}} \\
\cmidrule(lr){2-3} \cmidrule(lr){4-5} \cmidrule(lr){6-7}
 & \multicolumn{1}{c}{\emph{FQ}} & \multicolumn{1}{c|}{\emph{MU}} & \multicolumn{1}{c}{\emph{FQ}} & \multicolumn{1}{c|}{\emph{MU}} & \multicolumn{1}{c}{\emph{FQ}} & \multicolumn{1}{c}{\emph{MU}} \\

\midrule
Original & 1e-3  & 0.62  & 3e-16  & 0.62  & 2e-19  & 0.62  \\
Retrain & 1.0  & 0.62  & 1.0  & 0.62  & 1.0  & 0.62  \\
\midrule
GA & 0.57  & 0.55   & 0.05   & 0.02  & 8e-6  & 0  \\
GA+GD & 0.40  & 0.53 & 0.04  & 0.43  & 3e-6  & 0.48  \\
GA+KL & 0.05  & 0.56  & 6e-3  & 0.40  & 1e-5  & 0.33 \\
\midrule
NPO  & 0.71  & 0.56  & 0.54  & 0.15  & 0.1 & 0.07  \\
DPO+GD & 0.27  & 0.58  & 1e-4  & 0.02 & 5e-7  & 0.05  \\
NPO+GD & 0.71 & 0.58  & 0.74 & 0.53 & 0.45  & 0.55  \\
\midrule
\textbf{DiPO (ours)} & \textbf{0.99}  & \textbf{0.59} & \textbf{0.95}& \textbf{0.56} & \textbf{0.86} & \textbf{0.57}  \\
\bottomrule
\end{tabular}
\vspace{-1.5em}
\end{wraptable}
We first evaluate on the TOFU benchmark, which provides three levels of unlearning tasks (\text{TOFU-1\%}, \text{TOFU-5\%}, \text{TOFU-10\%}). The primary metrics include \emph{Forget Quality (FQ)}, measuring the extent of forgetting, and \emph{Model Utility (MU)}, evaluating model performance on the retain set. Detailed descriptions of the TOFU dataset, its evaluation metrics, and our hyperparameter settings are provided in \Cref{app:tofu}.

\paragraph{Effectiveness} As presented in \Cref{tab:tofu_results_best} (the ``best epoch'' refers to the training epoch that achieved the highest FQ), DiPO consistently achieves the best trade-off between FQ and MU compared to other optimization-based methods. For instance, on the \text{TOFU-10\%} task, DiPO improves FQ by over 20\% compared to the \text{NPO+GD} baseline while also exhibiting comparable MU. \Cref{figure:tofu_fq}(a) further illustrates this, showing \emph{DiPO positioned closest to the ideal ``Retrain LLM'' target}, particularly excelling in FQ. Notably, DiPO also achieves leading performance when considering the final epoch results (detailed comparison is in \Cref{tab:tofu_results_final}).
The examples presented in \Cref{tab:sample_table} further demonstrate DiPO's ability to achieve targeted forgetting while preserving accuracy on unrelated queries.

\paragraph{Training Stability} A significant advantage of DiPO is its training stability. As illustrated in \Cref{figure:tofu_fq}(b), DiPO maintains a stable, near-peak FQ value throughout the latter half of training, with its MU exhibiting a controlled adjustment before stabilizing. This contrasts with several baselines that show FQ declining after an initial peak and require early stopping to achieve optimal reported results. DiPO's consistent performance at the final epoch (detailed in \Cref{tab:tofu_results_final}) \emph{mitigates the need for such fragile early stopping}, enhancing its practical applicability.

\begin{wraptable}{r}{0.5\textwidth}
\vspace{-1.5em}
\centering
\caption{The best-epoch performance on TOFU benchmark among other unlearning framework. Scores closer to ``Retrain'' are better. \textbf{Bold} indicates best results among all methods.}
\label{tab:tofu_other_results_best}
\small 
\setlength{\tabcolsep}{3pt} 
\begin{tabular}{@{}c|cc|cc|cc@{}}
\toprule
\multicolumn{1}{c|}{\multirow{2}{*}{\textbf{Method}}} & \multicolumn{2}{c|}{\textbf{TOFU-1\%}} & \multicolumn{2}{c|}{\textbf{TOFU-5\%}} & \multicolumn{2}{c}{\textbf{TOFU-10\%}} \\
\cmidrule(lr){2-3} \cmidrule(lr){4-5} \cmidrule(lr){6-7}
 & \multicolumn{1}{c}{\emph{FQ}} & \multicolumn{1}{c|}{\emph{MU}} & \multicolumn{1}{c}{\emph{FQ}} & \multicolumn{1}{c|}{\emph{MU}} & \multicolumn{1}{c}{\emph{FQ}} & \multicolumn{1}{c}{\emph{MU}} \\

\midrule
Original & 1e-3  & 0.62  & 3e-16  & 0.62  & 2e-19  & 0.62  \\
Retrain & 1.0  & 0.62  & 1.0  & 0.62  & 1.0  & 0.62  \\
\midrule
ULD & \textbf{0.99}  & \textbf{0.62}  & 0.73  & \textbf{0.62}  & 0.48  & \textbf{0.62}  \\
AltPO & 0.92  & 0.55 & 0.71  & 0.54 & 0.58  & 0.56  \\
\midrule
\textbf{DiPO (ours)} & \textbf{0.99}  & 0.59 & \textbf{0.95}& 0.56 & \textbf{0.86} & 0.57  \\
\bottomrule
\end{tabular}
\end{wraptable}

\paragraph{Comparison with Other Frameworks}
We also compare DiPO with ULD and AltPO on TOFU. For ULD, while an open-source implementation is provided, our attempts to reproduce the published results did not yield comparable performance. Consequently, we refer to the results stated in the original work for our comparative analysis. For the AltPO and our method, we ran experiments with five random seeds and report the results from the best-performing seed. It is noteworthy that these methods employ \emph{TOFU-specific} data augmentation or auxiliary models (see \Cref{app:other_unlearn_frameworks}), intuitively granting them an advantage.
Nevertheless, \Cref{tab:tofu_other_results_best} shows DiPO achieves a markedly higher FQ value, surpassing AltPO by 48\% (0.86 vs. 0.58) and ULD by 79\% (0.86 vs. 0.48) on TOFU-10\%, \emph{without any additional components}. Instead, the ULD method uses the auxiliary model to prevent the erosion of retained knowledge and thus achieves high MU value. This significantly highlights DiPO's efficiency and potential for broader practical deployment due to its generalizability.


\subsection{Experiments on MUSE}
\label{sec:muse}
\begin{wraptable}{r}{0.5\textwidth}
\setlength{\tabcolsep}{3pt}
\vspace{-1.6em}
    \centering 
    \caption{Performance on MUSE. Scores closer to ``Retrain'' are better. Best results are in \textbf{bold}.} 
    \label{tab:muse_results}
    \vspace{3pt}
    \resizebox{\linewidth}{!}{%
        \begin{tabular}{@{}c | ccc | c@{}}
        \toprule
        \multicolumn{1}{c|}{\multirow{3}{*}{\textbf{Method}}} & \multicolumn{3}{c|}{\textbf{Unlearning Efficacy}} & \multicolumn{1}{c}{\textbf{Utility}} \\
        \cmidrule(lr){2-4} \cmidrule(lr){5-5}
         & VM-f & KM-f & PL($\rightarrow0$) & KM-r    \\
        \midrule
        Original & 58.3 & 62.9 & -99.8 & 54.3  \\
        Retrain  & 20.8 & 33.1 & 0.0 & 53.78  \\
        \midrule
        GA       & 0.0 & 0.0 & \textbf{5.2} & 0.0 \\
        GA+GD & 4.9 & \textbf{31.3} & 108.1 & 28.2  \\
        NPO      & 0.0 & 0.0 & 24.4 & 0.0  \\
        NPO+GD   & 1.2 & 54.6 & 105.8 & 40.5  \\
        \midrule
        \textbf{DiPO (ours)} & \textbf{31.67} & 53.22 & 98.1 & \textbf{51.46}  \\
        \bottomrule
        \end{tabular}%
    }
    \vspace{-0.9em}
\end{wraptable}
To further evaluate DiPO's generalization, we experiment on the BBC News corpus within MUSE, a recent and comprehensive benchmark of unlearning. MUSE employs multiple metrics, including \emph{VerbMem-f (VM-f)}, \emph{KnowMem-f (KM-f)}, and \emph{PrivLeak (PL)} for unlearning efficacy, \emph{KnowMem-r (KM-r)} for utility. It also includes \emph{Scalability} and \emph{Sustainability} to assess performance under increasing forget set sizes and sequential unlearning requests, respectively. More detailed descriptions and hyperparameter settings are provided in \Cref{app:muse}. Due to the TOFU-specific tailoring of ULD and AltPO, our MUSE comparisons only focus on optimization-based methods.

\paragraph{Results} As shown in \Cref{tab:muse_results}, DiPO demonstrates strong performance, achieving the best scores on VM-f and KM-r, which indicates effective verbatim unlearning and good knowledge retention, respectively. Furthermore, DiPO exhibits excellent Scalability and Sustainability in \Cref{figure:muse-add}(a), maintaining robust utility preservation as the forget set size increases (Scalability, left) and across sequential unlearning requests (Sustainability, right), outperforming baselines in dynamic scenarios. This underscores DiPO's potential for practical, large-scale applications.

\subsection{Additional analysis}
\label{additional-analysis}
\begin{figure}[t] 
    \centering
    \includegraphics[width=0.99\textwidth]{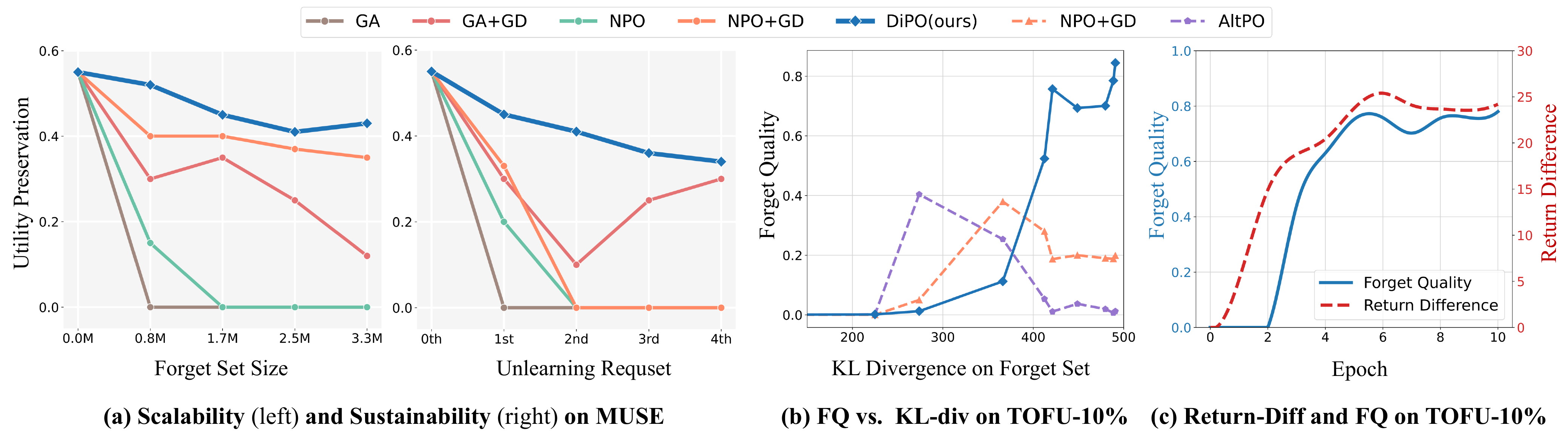}
    \caption{Robustness analysis on MUSE and DiPO's internal mechanisms. (a) Scalability and Sustainability performance on MUSE News. (b) FQ vs. KL Divergence on TOFU-10\% (from $\pi_{\text{ref}}$ on $\mathcal{D}_f$), demonstrating DiPO's higher unlearning efficiency. (c) Return Difference and FQ on TOFU-10\%, illustrating the correlation between DiPO's learned reward signals and unlearning efficacy.}
    \label{figure:muse-add}
    \vspace{-1em}
\end{figure}

In this section, we conduct further analyses on the \text{TOFU-10\%} settings and ablation studies on the whole TOFU benchmark, to provide deeper insights into DiPO's intrinsic mechanisms. 
\paragraph{Meaningful Deviation of KL Divergence}
We investigate how effectively DiPO converts the model divergence from $\pi_{\text{ref}}$ on $\mathcal{D}_f$ into unlearning, compared to baselines. \Cref{figure:muse-add}(b) plots FQ against KL divergence on TOFU-10\%. 
DiPO exhibits improved unlearning efficiency, with FQ substantially increasing even at higher KL values, indicating its updates are more ``targeted''. In contrast, NPO+GD shows FQ plateauing after an initial rise, suggesting its induced model changes are less effective for unlearning at higher divergences.  
Even AltPO, despite its engineered preferred responses, may exhibit lower efficiency in this regard compared to DiPO’s distribution-level manipulation. This supports that DiPO offers a more direct and efficient unlearning path.

\paragraph{Verification of DiPO's Reward Mechanism}
To empirically validate that DiPO's learning process aligns with its theoretical formulation (more details in \Cref{subsec:dipo_derivation}), we inspect the evolution of its internal distribution-level returns (specifically the difference between the preferred return $R_{\pi_w}$ and dispreferred return $R_{\pi_l}$) for the forget objective, plotted alongside FQ progression during training (\Cref{figure:muse-add}(c)). The widening gap between these returns, signifying better unlearning preference, strongly correlates with the improvement in FQ, particularly where rapid increases in the return difference align with significant FQ gains. This confirms that the learned preference signals effectively guide model unlearning.

\begin{wraptable}{r}{0.5\textwidth}
\vspace{-1.5em}
\centering
\caption{Ablation results. The value of each metric is averaged over five seeds at the best epoch. Best results are in \textbf{bold}.}
\vspace{0.25em}
\label{tab:tofu_results_ablation}
\small 
\setlength{\tabcolsep}{3pt} 
\begin{tabular}{@{}c|cc|cc|cc@{}}
\toprule
\multicolumn{1}{c|}{\multirow{2}{*}{\textbf{Method}}} & \multicolumn{2}{c|}{\textbf{TOFU-1\%}} & \multicolumn{2}{c|}{\textbf{TOFU-5\%}} & \multicolumn{2}{c}{\textbf{TOFU-10\%}} \\
\cmidrule(lr){2-3} \cmidrule(lr){4-5} \cmidrule(lr){6-7}
 & \multicolumn{1}{c}{\emph{FQ}} & \multicolumn{1}{c|}{\emph{MU}} & \multicolumn{1}{c}{\emph{FQ}} & \multicolumn{1}{c|}{\emph{MU}} & \multicolumn{1}{c}{\emph{FQ}} & \multicolumn{1}{c}{\emph{MU}} \\

\midrule
Original & 1e-3  & 0.62  & 3e-16  & 0.62  & 2e-19  & 0.62  \\
Retrain & 1.0  & 0.62  & 1.0  & 0.62  & 1.0  & 0.62  \\
\midrule
\textbf{DiPO (ours)} & \textbf{0.89} & 0.58 & \textbf{0.95} & 0.58 & \textbf{0.84} & 0.56 \\
DiPO(f)+GD & 0.57  & \textbf{0.62} & 0.54  & \textbf{0.62}  & 3e-5  & \textbf{0.65}  \\
GA+DiPO(r) & 0.16  & 0.39  & 1e-13  & 0.59  & 3e-10  & 0.38 \\
NPO+DiPO(r) & 0.12  & 0.55  & 0.07  & 0.01  & 3e-2 & 4e-3  \\

\bottomrule
\end{tabular}
\vspace{-0.5em}
\end{wraptable}

\paragraph{Ablation Studies}

We investigate the interplay of DiPO's core $\mathcal{L}_{\text{DiPO-f}}$ and $\mathcal{L}_{\text{DiPO-r}}$ in \Cref{tab:tofu_results_ablation}. Our main DiPO (using both $\mathcal{L}_{\text{DiPO-f}}$ and $\mathcal{L}_{\text{DiPO-r}}$) is compared against variants where one DiPO component is substituted with another loss, detailed in \Cref{app:ablation-studies}.
The results compellingly show that while $\mathcal{L}_{\text{GD}}$ can significantly boost MU, the effective trade-off between FQ and MU is achieved only with our main DiPO configuration. This underscores that DiPO's strength lies in its integrated, preference-based design for both forget and retain objectives.
Furthermore, as detailed in \Cref{figure:ablation-curves}, we analyze the performance of using only $\mathcal{L}_{\text{DiPO-f}}$, and find it achieves effective unlearning while maintaining a degree of MU. This is a significant advantage over typical baselines relying solely on forget loss (such as GA and NPO), which tend to exhibit a collapse in both metrics. This finding highlights the inherent robustness and targeted nature of the DiPO forget mechanism itself, even in the absence of an explicit retain objective.

\section{Conclusion}
In this paper, we propose the distribution-level for LLM unlearning, a fine-grained perspective which can overcome the limitations of response-level approaches. Building upon this, we derive a novel algorithm, \textbf{Di}stribution \textbf{P}reference \textbf{O}ptimization (DiPO), along with an intrinsic method for constructing complete preference distribution pairs directly from model logits. This provides precise guidance for the unlearning process without requiring auxiliary models or domain-specific knowledge, thereby enhancing its generalizability. Both theoretical analysis and extensive experimental results demonstrate the effectiveness and stability of our method.



\bibliographystyle{unsrtnat}
\bibliography{main}

\begin{thebibliography}{54}
\providecommand{\natexlab}[1]{#1}
\providecommand{\url}[1]{\texttt{#1}}
\expandafter\ifx\csname urlstyle\endcsname\relax
  \providecommand{\doi}[1]{doi: #1}\else
  \providecommand{\doi}{doi: \begingroup \urlstyle{rm}\Url}\fi

\bibitem[Yu et~al.(2023)Yu, Jeoung, Kasi, Yu, and Ji]{yu2023unlearning}
Charles Yu, Sullam Jeoung, Anish Kasi, Pengfei Yu, and Heng Ji.
\newblock Unlearning bias in language models by partitioning gradients.
\newblock In \emph{Findings of the Association for Computational Linguistics: ACL 2023}, pages 6032--6048, 2023.

\bibitem[Liu et~al.(2022)Liu, Liu, and Stone]{liu2022continual}
Bo~Liu, Qiang Liu, and Peter Stone.
\newblock Continual learning and private unlearning.
\newblock In \emph{Conference on Lifelong Learning Agents}, pages 243--254. PMLR, 2022.

\bibitem[Eldan and Russinovich(2023)]{eldan2023s}
Ronen Eldan and Mark Russinovich.
\newblock Who’s harry potter? approximate unlearning for llms.
\newblock 2023.

\bibitem[Jang et~al.(2022)Jang, Yoon, Yang, Cha, Lee, Logeswaran, and Seo]{jang2022knowledge}
Joel Jang, Dongkeun Yoon, Sohee Yang, Sungmin Cha, Moontae Lee, Lajanugen Logeswaran, and Minjoon Seo.
\newblock Knowledge unlearning for mitigating privacy risks in language models.
\newblock \emph{arXiv preprint arXiv:2210.01504}, 2022.

\bibitem[Wu et~al.(2023{\natexlab{a}})Wu, Li, Xu, Dong, Wu, Bian, and Xiong]{wu2023depn}
Xinwei Wu, Junzhuo Li, Minghui Xu, Weilong Dong, Shuangzhi Wu, Chao Bian, and Deyi Xiong.
\newblock Depn: Detecting and editing privacy neurons in pretrained language models.
\newblock \emph{arXiv preprint arXiv:2310.20138}, 2023{\natexlab{a}}.

\bibitem[Liu et~al.(2024{\natexlab{a}})Liu, Dou, Tan, Tian, and Jiang]{liu2024towards}
Zheyuan Liu, Guangyao Dou, Zhaoxuan Tan, Yijun Tian, and Meng Jiang.
\newblock Towards safer large language models through machine unlearning.
\newblock \emph{arXiv preprint arXiv:2402.10058}, 2024{\natexlab{a}}.

\bibitem[Barrett et~al.(2023)Barrett, Boyd, Bursztein, Carlini, Chen, Choi, Chowdhury, Christodorescu, Datta, Feizi, et~al.]{barrett2023identifying}
Clark Barrett, Brad Boyd, Elie Bursztein, Nicholas Carlini, Brad Chen, Jihye Choi, Amrita~Roy Chowdhury, Mihai Christodorescu, Anupam Datta, Soheil Feizi, et~al.
\newblock Identifying and mitigating the security risks of generative ai.
\newblock \emph{Foundations and Trends{\textregistered} in Privacy and Security}, 6\penalty0 (1):\penalty0 1--52, 2023.

\bibitem[{European Parliament and Council of the European Union}(2016)]{GDPR2016}
{European Parliament and Council of the European Union}.
\newblock Regulation ({EU}) 2016/679 of the {European Parliament} and of the {Council} of 27 april 2016 on the protection of natural persons with regard to the processing of personal data and on the free movement of such data, and repealing directive 95/46/{EC} ({General Data Protection Regulation}).
\newblock Official Journal of the European Union, April 2016.
\newblock Published in OJ L 119, 4.5.2016, pp. 1--88. Adopted 27 April 2016.

\bibitem[{California State Assembly}(2018)]{CCPA_AB375_2018}
{California State Assembly}.
\newblock Assembly bill no. 375 ({Chapter} 55, statutes of 2018). an act to add title 1.81.5 (commencing with section 1798.100) to part 4 of division 3 of the civil code, relating to privacy. ({California Consumer Privacy Act of 2018}).
\newblock California Legislature, 2017--2018 Regular Session, June 2018.
\newblock Approved by Governor and filed with Secretary of State June 28, 2018. This bill enacted the CCPA.

\bibitem[Yao et~al.(2024{\natexlab{a}})Yao, Xu, and Liu]{yao2024large}
Yuanshun Yao, Xiaojun Xu, and Yang Liu.
\newblock Large language model unlearning.
\newblock \emph{Advances in Neural Information Processing Systems}, 37:\penalty0 105425--105475, 2024{\natexlab{a}}.

\bibitem[Zhang et~al.(2024{\natexlab{a}})Zhang, Lin, Bai, and Mei]{zhang2024negative}
Ruiqi Zhang, Licong Lin, Yu~Bai, and Song Mei.
\newblock Negative preference optimization: From catastrophic collapse to effective unlearning.
\newblock In \emph{First Conference on Language Modeling}, 2024{\natexlab{a}}.

\bibitem[Rafailov et~al.(2023)Rafailov, Sharma, Mitchell, Manning, Ermon, and Finn]{rafailov2023direct}
Rafael Rafailov, Archit Sharma, Eric Mitchell, Christopher~D Manning, Stefano Ermon, and Chelsea Finn.
\newblock Direct preference optimization: Your language model is secretly a reward model.
\newblock \emph{Advances in Neural Information Processing Systems}, 36:\penalty0 53728--53741, 2023.

\bibitem[Maini et~al.(2024)Maini, Feng, Schwarzschild, Lipton, and Kolter]{maini2024tofu}
Pratyush Maini, Zhili Feng, Avi Schwarzschild, Zachary~Chase Lipton, and J~Zico Kolter.
\newblock {TOFU}: A task of fictitious unlearning for {LLM}s.
\newblock In \emph{First Conference on Language Modeling}, 2024.

\bibitem[Shi et~al.(2025)Shi, Lee, Huang, Malladi, Zhao, Holtzman, Liu, Zettlemoyer, Smith, and Zhang]{shi2025muse}
Weijia Shi, Jaechan Lee, Yangsibo Huang, Sadhika Malladi, Jieyu Zhao, Ari Holtzman, Daogao Liu, Luke Zettlemoyer, Noah~A. Smith, and Chiyuan Zhang.
\newblock {MUSE}: Machine unlearning six-way evaluation for language models.
\newblock In \emph{The Thirteenth International Conference on Learning Representations}, 2025.

\bibitem[Nguyen et~al.(2022)Nguyen, Huynh, Ren, Nguyen, Liew, Yin, and Nguyen]{nguyen2022survey}
Thanh~Tam Nguyen, Thanh~Trung Huynh, Zhao Ren, Phi~Le Nguyen, Alan Wee-Chung Liew, Hongzhi Yin, and Quoc Viet~Hung Nguyen.
\newblock A survey of machine unlearning.
\newblock \emph{arXiv preprint arXiv:2209.02299}, 2022.

\bibitem[Cao and Yang(2015)]{cao2015towards}
Yinzhi Cao and Junfeng Yang.
\newblock Towards making systems forget with machine unlearning.
\newblock In \emph{2015 IEEE symposium on security and privacy}, pages 463--480. IEEE, 2015.

\bibitem[Thudi et~al.(2022)Thudi, Deza, Chandrasekaran, and Papernot]{thudi2022unrolling}
Anvith Thudi, Gabriel Deza, Varun Chandrasekaran, and Nicolas Papernot.
\newblock Unrolling sgd: Understanding factors influencing machine unlearning.
\newblock In \emph{2022 IEEE 7th European Symposium on Security and Privacy (EuroS\&P)}, pages 303--319. IEEE, 2022.

\bibitem[Izzo et~al.(2021)Izzo, Smart, Chaudhuri, and Zou]{izzo2021approximate}
Zachary Izzo, Mary~Anne Smart, Kamalika Chaudhuri, and James Zou.
\newblock Approximate data deletion from machine learning models.
\newblock In \emph{International Conference on Artificial Intelligence and Statistics}, pages 2008--2016. PMLR, 2021.

\bibitem[Wang et~al.(2023{\natexlab{a}})Wang, Chen, Yuan, Zeng, Wong, and Yin]{wang2023kga}
Lingzhi Wang, Tong Chen, Wei Yuan, Xingshan Zeng, Kam-Fai Wong, and Hongzhi Yin.
\newblock Kga: A general machine unlearning framework based on knowledge gap alignment.
\newblock \emph{arXiv preprint arXiv:2305.06535}, 2023{\natexlab{a}}.

\bibitem[Triantafillou et~al.(2024)Triantafillou, Kairouz, Pedregosa, Hayes, Kurmanji, Zhao, Dumoulin, Junior, Mitliagkas, Wan, et~al.]{triantafillou2024we}
Eleni Triantafillou, Peter Kairouz, Fabian Pedregosa, Jamie Hayes, Meghdad Kurmanji, Kairan Zhao, Vincent Dumoulin, Julio~Jacques Junior, Ioannis Mitliagkas, Jun Wan, et~al.
\newblock Are we making progress in unlearning? findings from the first neurips unlearning competition.
\newblock \emph{arXiv preprint arXiv:2406.09073}, 2024.

\bibitem[Golatkar et~al.(2020)Golatkar, Achille, and Soatto]{golatkar2020eternal}
Aditya Golatkar, Alessandro Achille, and Stefano Soatto.
\newblock Eternal sunshine of the spotless net: Selective forgetting in deep networks.
\newblock In \emph{Proceedings of the IEEE/CVF conference on computer vision and pattern recognition}, pages 9304--9312, 2020.

\bibitem[Bourtoule et~al.(2021)Bourtoule, Chandrasekaran, Choquette-Choo, Jia, Travers, Zhang, Lie, and Papernot]{bourtoule2021machine}
Lucas Bourtoule, Varun Chandrasekaran, Christopher~A Choquette-Choo, Hengrui Jia, Adelin Travers, Baiwu Zhang, David Lie, and Nicolas Papernot.
\newblock Machine unlearning.
\newblock In \emph{2021 IEEE symposium on security and privacy (SP)}, pages 141--159. IEEE, 2021.

\bibitem[Jia et~al.(2023)Jia, Liu, Ram, Yao, Liu, Liu, Sharma, and Liu]{jia2023model}
Jinghan Jia, Jiancheng Liu, Parikshit Ram, Yuguang Yao, Gaowen Liu, Yang Liu, Pranay Sharma, and Sijia Liu.
\newblock Model sparsity can simplify machine unlearning.
\newblock \emph{Advances in Neural Information Processing Systems}, 36:\penalty0 51584--51605, 2023.

\bibitem[Fan et~al.(2023)Fan, Liu, Zhang, Wong, Wei, and Liu]{fan2023salun}
Chongyu Fan, Jiancheng Liu, Yihua Zhang, Eric Wong, Dennis Wei, and Sijia Liu.
\newblock Salun: Empowering machine unlearning via gradient-based weight saliency in both image classification and generation.
\newblock \emph{arXiv preprint arXiv:2310.12508}, 2023.

\bibitem[Kurmanji et~al.(2023)Kurmanji, Triantafillou, Hayes, and Triantafillou]{kurmanji2023towards}
Meghdad Kurmanji, Peter Triantafillou, Jamie Hayes, and Eleni Triantafillou.
\newblock Towards unbounded machine unlearning.
\newblock \emph{Advances in neural information processing systems}, 36:\penalty0 1957--1987, 2023.

\bibitem[Ginart et~al.(2019)Ginart, Guan, Valiant, and Zou]{ginart2019making}
Antonio Ginart, Melody Guan, Gregory Valiant, and James~Y Zou.
\newblock Making ai forget you: Data deletion in machine learning.
\newblock \emph{Advances in neural information processing systems}, 32, 2019.

\bibitem[Gandikota et~al.(2023)Gandikota, Materzynska, Fiotto-Kaufman, and Bau]{gandikota2023erasing}
Rohit Gandikota, Joanna Materzynska, Jaden Fiotto-Kaufman, and David Bau.
\newblock Erasing concepts from diffusion models.
\newblock In \emph{Proceedings of the IEEE/CVF International Conference on Computer Vision}, pages 2426--2436, 2023.

\bibitem[Zhang et~al.(2024{\natexlab{b}})Zhang, Wang, Xu, Wang, and Shi]{zhang2024forget}
Gong Zhang, Kai Wang, Xingqian Xu, Zhangyang Wang, and Humphrey Shi.
\newblock Forget-me-not: Learning to forget in text-to-image diffusion models.
\newblock In \emph{Proceedings of the IEEE/CVF conference on computer vision and pattern recognition}, pages 1755--1764, 2024{\natexlab{b}}.

\bibitem[Che et~al.(2023)Che, Zhou, Zhang, Lyu, Liu, Yan, Dou, and Huan]{che2023fast}
Tianshi Che, Yang Zhou, Zijie Zhang, Lingjuan Lyu, Ji~Liu, Da~Yan, Dejing Dou, and Jun Huan.
\newblock Fast federated machine unlearning with nonlinear functional theory.
\newblock In \emph{International conference on machine learning}, pages 4241--4268. PMLR, 2023.

\bibitem[Pan et~al.(2025)Pan, Wang, Li, Zheng, Wang, Tang, and Zhao]{pan2025federated}
Zibin Pan, Zhichao Wang, Chi Li, Kaiyan Zheng, Boqi Wang, Xiaoying Tang, and Junhua Zhao.
\newblock Federated unlearning with gradient descent and conflict mitigation.
\newblock In \emph{Proceedings of the AAAI Conference on Artificial Intelligence}, volume~39, pages 19804--19812, 2025.

\bibitem[Chien et~al.(2022)Chien, Pan, and Milenkovic]{chien2022efficient}
Eli Chien, Chao Pan, and Olgica Milenkovic.
\newblock Efficient model updates for approximate unlearning of graph-structured data.
\newblock In \emph{The Eleventh International Conference on Learning Representations}, 2022.

\bibitem[Wu et~al.(2023{\natexlab{b}})Wu, Shen, Ning, Wang, and Wang]{wu2023certified}
Kun Wu, Jie Shen, Yue Ning, Ting Wang, and Wendy~Hui Wang.
\newblock Certified edge unlearning for graph neural networks.
\newblock In \emph{Proceedings of the 29th ACM SIGKDD Conference on Knowledge Discovery and Data Mining}, pages 2606--2617, 2023{\natexlab{b}}.

\bibitem[Sachdeva et~al.(2024)Sachdeva, Rathee, Sristi, Sharma, and Wydma{\'n}ski]{sachdeva2024machine}
Bhavika Sachdeva, Harshita Rathee, Sristi, Arun Sharma, and Witold Wydma{\'n}ski.
\newblock Machine unlearning for recommendation systems: An insight.
\newblock In \emph{International Conference On Innovative Computing And Communication}, pages 415--430. Springer, 2024.

\bibitem[Fan et~al.(2024)Fan, Liu, Lin, Jia, Zhang, Mei, and Liu]{fan2024simplicity}
Chongyu Fan, Jiancheng Liu, Licong Lin, Jinghan Jia, Ruiqi Zhang, Song Mei, and Sijia Liu.
\newblock Simplicity prevails: Rethinking negative preference optimization for llm unlearning.
\newblock \emph{arXiv preprint arXiv:2410.07163}, 2024.

\bibitem[Mekala et~al.(2024)Mekala, Dorna, Dubey, Lalwani, Koleczek, Rungta, Hasan, and Lobo]{mekala2024alternate}
Anmol Mekala, Vineeth Dorna, Shreya Dubey, Abhishek Lalwani, David Koleczek, Mukund Rungta, Sadid Hasan, and Elita Lobo.
\newblock Alternate preference optimization for unlearning factual knowledge in large language models.
\newblock \emph{arXiv preprint arXiv:2409.13474}, 2024.

\bibitem[Jia et~al.(2024)Jia, Zhang, Zhang, Liu, Runwal, Diffenderfer, Kailkhura, and Liu]{jia2024soul}
Jinghan Jia, Yihua Zhang, Yimeng Zhang, Jiancheng Liu, Bharat Runwal, James Diffenderfer, Bhavya Kailkhura, and Sijia Liu.
\newblock Soul: Unlocking the power of second-order optimization for llm unlearning.
\newblock \emph{arXiv preprint arXiv:2404.18239}, 2024.

\bibitem[Chundawat et~al.(2023)Chundawat, Tarun, Mandal, and Kankanhalli]{chundawat2023can}
Vikram~S Chundawat, Ayush~K Tarun, Murari Mandal, and Mohan Kankanhalli.
\newblock Can bad teaching induce forgetting? unlearning in deep networks using an incompetent teacher.
\newblock In \emph{Proceedings of the AAAI Conference on Artificial Intelligence}, volume~37, pages 7210--7217, 2023.

\bibitem[Ji et~al.(2024)Ji, Liu, Zhang, Liu, Kompella, Liu, and Chang]{ji2024reversing}
Jiabao Ji, Yujian Liu, Yang Zhang, Gaowen Liu, Ramana Kompella, Sijia Liu, and Shiyu Chang.
\newblock Reversing the forget-retain objectives: An efficient llm unlearning framework from logit difference.
\newblock \emph{Advances in Neural Information Processing Systems}, 37:\penalty0 12581--12611, 2024.

\bibitem[Chen and Yang(2023)]{chen2023unlearn}
Jiaao Chen and Diyi Yang.
\newblock Unlearn what you want to forget: Efficient unlearning for llms.
\newblock \emph{arXiv preprint arXiv:2310.20150}, 2023.

\bibitem[Yao et~al.(2024{\natexlab{b}})Yao, Chien, Du, Niu, Wang, Cheng, and Yue]{yao2024machine}
Jin Yao, Eli Chien, Minxin Du, Xinyao Niu, Tianhao Wang, Zezhou Cheng, and Xiang Yue.
\newblock Machine unlearning of pre-trained large language models.
\newblock \emph{arXiv preprint arXiv:2402.15159}, 2024{\natexlab{b}}.

\bibitem[Ishibashi and Shimodaira(2023)]{ishibashi2023knowledge}
Yoichi Ishibashi and Hidetoshi Shimodaira.
\newblock Knowledge sanitization of large language models.
\newblock \emph{arXiv preprint arXiv:2309.11852}, 2023.

\bibitem[Liu et~al.(2024{\natexlab{b}})Liu, Zhang, Jaakkola, and Chang]{liu2024revisiting}
Yujian Liu, Yang Zhang, Tommi Jaakkola, and Shiyu Chang.
\newblock Revisiting who's harry potter: Towards targeted unlearning from a causal intervention perspective.
\newblock \emph{arXiv preprint arXiv:2407.16997}, 2024{\natexlab{b}}.

\bibitem[Thaker et~al.(2024)Thaker, Maurya, Hu, Wu, and Smith]{thaker2024guardrail}
Pratiksha Thaker, Yash Maurya, Shengyuan Hu, Zhiwei~Steven Wu, and Virginia Smith.
\newblock Guardrail baselines for unlearning in llms.
\newblock \emph{arXiv preprint arXiv:2403.03329}, 2024.

\bibitem[Pawelczyk et~al.(2023)Pawelczyk, Neel, and Lakkaraju]{pawelczyk2023context}
Martin Pawelczyk, Seth Neel, and Himabindu Lakkaraju.
\newblock In-context unlearning: Language models as few shot unlearners.
\newblock \emph{arXiv preprint arXiv:2310.07579}, 2023.

\bibitem[Ouyang et~al.(2022)Ouyang, Wu, Jiang, Almeida, Wainwright, Mishkin, Zhang, Agarwal, Slama, Ray, et~al.]{ouyang2022training}
Long Ouyang, Jeffrey Wu, Xu~Jiang, Diogo Almeida, Carroll Wainwright, Pamela Mishkin, Chong Zhang, Sandhini Agarwal, Katarina Slama, Alex Ray, et~al.
\newblock Training language models to follow instructions with human feedback.
\newblock \emph{Advances in neural information processing systems}, 35:\penalty0 27730--27744, 2022.

\bibitem[Hong et~al.(2024)Hong, Lee, and Thorne]{hong2024orpo}
Jiwoo Hong, Noah Lee, and James Thorne.
\newblock Orpo: Monolithic preference optimization without reference model.
\newblock \emph{arXiv preprint arXiv:2403.07691}, 2024.

\bibitem[Meng et~al.(2024)Meng, Xia, and Chen]{meng2024simpo}
Yu~Meng, Mengzhou Xia, and Danqi Chen.
\newblock Simpo: Simple preference optimization with a reference-free reward.
\newblock \emph{Advances in Neural Information Processing Systems}, 37:\penalty0 124198--124235, 2024.

\bibitem[Wang et~al.(2023{\natexlab{b}})Wang, Jiang, Yang, Liu, and Chen]{wang2023beyond}
Chaoqi Wang, Yibo Jiang, Chenghao Yang, Han Liu, and Yuxin Chen.
\newblock Beyond reverse kl: Generalizing direct preference optimization with diverse divergence constraints.
\newblock \emph{arXiv preprint arXiv:2309.16240}, 2023{\natexlab{b}}.

\bibitem[Azar et~al.(2024)Azar, Guo, Piot, Munos, Rowland, Valko, and Calandriello]{azar2024general}
Mohammad~Gheshlaghi Azar, Zhaohan~Daniel Guo, Bilal Piot, Remi Munos, Mark Rowland, Michal Valko, and Daniele Calandriello.
\newblock A general theoretical paradigm to understand learning from human preferences.
\newblock In \emph{International Conference on Artificial Intelligence and Statistics}, pages 4447--4455. PMLR, 2024.

\bibitem[Sun et~al.(2024)Sun, Zheng, Zhao, Chang, and Wang]{sun2024generalizing}
Haoyuan Sun, Yuxin Zheng, Yifei Zhao, Yongzhe Chang, and Xueqian Wang.
\newblock Generalizing offline alignment theoretical paradigm with diverse divergence constraints.
\newblock In \emph{ICML 2024 Workshop on Models of Human Feedback for AI Alignment}, 2024.

\bibitem[Zeng et~al.(2024)Zeng, Liu, Ma, Yang, Zhang, and Wang]{zeng2024token}
Yongcheng Zeng, Guoqing Liu, Weiyu Ma, Ning Yang, Haifeng Zhang, and Jun Wang.
\newblock Token-level direct preference optimization.
\newblock \emph{arXiv preprint arXiv:2404.11999}, 2024.

\bibitem[Li et~al.(2022)Li, Holtzman, Fried, Liang, Eisner, Hashimoto, Zettlemoyer, and Lewis]{li2022contrastive}
Xiang~Lisa Li, Ari Holtzman, Daniel Fried, Percy Liang, Jason Eisner, Tatsunori Hashimoto, Luke Zettlemoyer, and Mike Lewis.
\newblock Contrastive decoding: Open-ended text generation as optimization.
\newblock \emph{arXiv preprint arXiv:2210.15097}, 2022.

\bibitem[Chuang et~al.(2023)Chuang, Xie, Luo, Kim, Glass, and He]{chuang2023dola}
Yung-Sung Chuang, Yujia Xie, Hongyin Luo, Yoon Kim, James Glass, and Pengcheng He.
\newblock Dola: Decoding by contrasting layers improves factuality in large language models.
\newblock \emph{arXiv preprint arXiv:2309.03883}, 2023.

\bibitem[Liu et~al.(2021)Liu, Sap, Lu, Swayamdipta, Bhagavatula, Smith, and Choi]{liu2021dexperts}
Alisa Liu, Maarten Sap, Ximing Lu, Swabha Swayamdipta, Chandra Bhagavatula, Noah~A Smith, and Yejin Choi.
\newblock Dexperts: Decoding-time controlled text generation with experts and anti-experts.
\newblock \emph{arXiv preprint arXiv:2105.03023}, 2021.

\end{thebibliography}

\appendix
\section{Limitations}
\label{app:limitation}
Despite DiPO demonstrating strong unlearning capabilities, certain limitations warrant discussion. First, similar to other current unlearning methods, DiPO's outputs are not entirely immune to hallucination, reflecting an ongoing challenge in the field. 
Second, while our intrinsic mechanism for constructing preference pairs is effective and general, its current simplicity may not fully address the complexities required for unlearning against information leakage, such as those evaluated by Membership Inference Attacks (MIAs). This is indicated by DiPO's performance on challenging privacy-related metrics, like the PrivLeak scores in the MUSE benchmark, where more sophisticated preference modeling might be beneficial. We plan to explore these problems in future work.

\section{Theoretical Details}
\label{app:theoretical-details}
\subsection{Distribution-level Return Derivation}
\label{app:Return derivation}
In \Cref{subsec:dipo_derivation} we showe the immediate reward function $r_\pi(x, y^{<t})$:
\begin{align*}
r_\pi(x, y^{<t}) &= \mathbb{E}_{z\sim\pi(\cdot|[x,y^{<t}])}[r([x, y^{<t}], z)] \nonumber\\
&= \mathbb{E}_{z\sim\pi(\cdot|[x,y^{<t}])}[\beta\log\frac{\pi_\theta^*(z|[x, y^{<t}])}{\pi_{\text{ref}}(z|[x, y^{<t}])} + \beta D_{KL}(\pi_{\text{ref}}(\cdot|[x,y^{<t}])||\pi_\theta^*(\cdot|[x,y^{<t}]))] \nonumber\\
&= \beta\mathbb{E}_{z\sim\pi(\cdot|[x,y^{<t}])}\left[\log\frac{\pi_\theta^*(z|[x, y^{<t}])}{\pi_{\text{ref}}(z|[x, y^{<t}])}\right] + \beta D_{KL}(\pi_{\text{ref}}(\cdot|[x,y^{<t}])||\pi_\theta^*(\cdot|[x,y^{<t}]))
\end{align*}
Using the definition of KL divergence, the expectation term can be rewritten as:
\begin{align*}
&\mathbb{E}_{z\sim\pi(\cdot|[x,y^{<t}])}\left[\log\frac{\pi_\theta^*(z|[x, y^{<t}])}{\pi_{\text{ref}}(z|[x, y^{<t}])}\right] \\
&= \mathbb{E}_{z\sim\pi(\cdot|[x,y^{<t}])}\left[\log\frac{\pi_\theta^*(z|[x, y^{<t}])}{\pi(z|[x, y^{<t}])} \cdot \frac{\pi(z|[x, y^{<t}])}{\pi_{\text{ref}}(z|[x, y^{<t}])}\right] \\
&= \mathbb{E}_{z\sim\pi(\cdot|[x,y^{<t}])}\left[\log\frac{\pi(z|[x, y^{<t}])}{\pi_{\text{ref}}(z|[x, y^{<t}])}\right] - \mathbb{E}_{z\sim\pi(\cdot|[x,y^{<t}])}\left[\log\frac{\pi(z|[x, y^{<t}])}{\pi_\theta^*(z|[x, y^{<t}])}\right] \\
&= D_{KL}(\pi(\cdot|[x,y^{<t}])||\pi_{\text{ref}}(\cdot|[x,y^{<t}])) - D_{KL}(\pi(\cdot|[x,y^{<t}])||\pi_\theta^*(\cdot|[x,y^{<t}])).
\end{align*}
For a response $y$ (i.e. a specific trajectory in RL), we can calculate the return $R_\pi(x, y)$  as follows:
\begin{align*}
    R_\pi(x, y) &= \sum_{t=1}^T r_\pi(x, y^{<t})\\
    &= \sum_{t=1}^T \beta D_{KL}(\pi(\cdot|[x,y^{<t}])||\pi_{\text{ref}}(\cdot|[x,y^{<t}])) \\
    & \quad - \beta \sum_{t=1}^T D_{KL}(\pi(\cdot|[x,y^{<t}])||\pi_\theta^*(\cdot|[x,y^{<t}])) + \sum_{t=1}^T \beta D_{KL}(\pi_{\text{ref}}(\cdot|[x,y^{<t}])||\pi_\theta^*(\cdot|[x,y^{<t}])).
\end{align*}
This is the formula in \Cref{eq:dipo-returns}.

\subsection{Detailed proof of DiPO loss}
\label{app:dipo-loss-proof}
Recall from \Cref{eq:dipo-returns} that the distribution-level return is:
\begin{equation}
    R_\pi(x, y, \pi_\theta^*) \coloneqq R_\pi(x, y) = \beta D_{SeqKL}(x, y; \pi||\pi_{\text{ref}}) - \beta D_{SeqKL}(x, y; \pi||\pi_\theta^*) + \beta D_{SeqKL}(x, y; \pi_{\text{ref}}||\pi_\theta^*)\nonumber.
\end{equation}
Given a specific sample $(x,y)$ and a pair of preference distributions $(\pi_w,\pi_l)$, we can derive their respective return expressions:
\begin{align}
    R_{\pi_w}(x, y, \pi_\theta^*) &= \beta D_{SeqKL}(x, y; {\pi_w}||\pi_{\text{ref}}) - \beta D_{SeqKL}(x, y; {\pi_w}||\pi_\theta^*) + \beta D_{SeqKL}(x, y; \pi_{\text{ref}}||\pi_\theta^*), \\
    R_{\pi_l}(x, y, \pi_\theta^*) &= \beta D_{SeqKL}(x, y; {\pi_l}||\pi_{\text{ref}}) - \beta D_{SeqKL}(x, y; {\pi_l}||\pi_\theta^*) + \beta D_{SeqKL}(x, y; \pi_{\text{ref}}||\pi_\theta^*).
\end{align}
These respectively represent the degree of preference for response $y$ under different policies. Consequently, we can employ BT model to construct the preference model:
\begin{align}
    p^*(R_{\pi_w} \succ R_{\pi_l}|(x,y)) &= \frac{\exp(R_{\pi_w}(x, y, \pi_\theta^*))}{\exp(R_{\pi_w}(x, y, \pi_\theta^*)) + \exp(R_{\pi_l}(x, y, \pi_\theta^*))} \nonumber\\
    &= \frac{1}{1 + \exp(R_{\pi_l}(x, y, \pi_\theta^*)-R_{\pi_w}(x, y, \pi_\theta^*))}.
\end{align}
Now that we have the probability of human preference data in terms of the optimal policy rather than the reward model, we can formulate a maximum likelihood objective for a parametrized policy $\pi_\theta$. Similar to the DPO method, our policy objective becomes:
\begin{align}
&\mathcal{L}_{\text{DiPO}}(\pi_\theta;\pi_w,\pi_l,\pi_{\text{ref}}) \nonumber\\
&= -\mathbb{E}_{(x,y)\sim\mathcal{D}}\left[\log p(R_{\pi_w} \succ R_{\pi_l}|(x,y)) \right] \nonumber\\
&= -\mathbb{E}_{(x,y)\sim\mathcal{D}}\left[\log \frac{1}{1 + \exp(R_{\pi_l}(x, y, \pi_\theta)-R_{\pi_w}(x, y, \pi_\theta))} \right] \nonumber\\
&= -\mathbb{E}_{(x,y)\sim\mathcal{D}}\left[\log\sigma\left(  \Big( R_{\pi_w}(x, y, \pi_\theta)-R_{\pi_l}(x, y, \pi_\theta) \Big) \right) \right] \nonumber\\
&= -\mathbb{E}_{(x,y)\sim\mathcal{D}}\left[\log\sigma\left(  \Big( \beta D_{SeqKL}(x, y; {\pi_w}||\pi_{\text{ref}}) - \beta D_{SeqKL}(x, y; {\pi_w}||\pi_\theta^*) + \beta D_{SeqKL}(x, y; \pi_{\text{ref}}||\pi_\theta^*) \right. \right. \nonumber \\
& \qquad \left. \left. - \Big(\beta D_{SeqKL}(x, y; {\pi_l}||\pi_{\text{ref}}) - \beta D_{SeqKL}(x, y; {\pi_l}||\pi_\theta^*) + \beta D_{SeqKL}(x, y; \pi_{\text{ref}}||\pi_\theta^*)\Big) \Big) \right) \right] \nonumber\\
&= -\mathbb{E}_{(x,y)\sim\mathcal{D}}\left[\log\sigma\left( \beta \Big( D_{SeqKL}(x, y; \pi_l||\pi_{\theta}) - D_{SeqKL}(x, y; \pi_w||\pi_{\theta}) \Big) \right. \right. \nonumber \\
& \qquad \left. \left. + \beta \Big( D_{SeqKL}(x, y; \pi_w||\pi_{\text{ref}}) - D_{SeqKL}(x, y; \pi_l||\pi_{\text{ref}}) \Big) \right) \right].
\end{align}
Now that we have the loss function of DiPO.

\subsection{Pseudo-code of DiPO}
\label{Pseudo-code}
\begin{algorithm}[h]
\caption{Distribution Preference Optimization (DiPO)}
\label{alg:dipo_pseudo_code}
\begin{algorithmic}[1] 
\State \textbf{Input:} Datasets $\mathcal{D}_f, \mathcal{D}_r$, Reference model $\pi_{\text{ref}}$, Policy model $\pi_\theta$, $\beta_f, \beta_r, \eta, \lambda, p$
\State \textbf{Initialize:} $\theta \leftarrow \theta_{\text{ref}}$
\For{each training epoch} 
    \State Sample mini-batches $B_f \sim \mathcal{D}_f$, $B_r \sim \mathcal{D}_r$
    \State Generate approx. $\pi_f(\cdot | x_f, y_f^{<t}), \pi_m(\cdot | x_f, y_f^{<t})$ from $\pi_\theta$ for $(x_f,y_f) \in B_f$ via top-$p$ logit filtering
    \State Generate approx. $\pi_f(\cdot | x_r, y_r^{<t}), \pi_m(\cdot | x_r, y_r^{<t})$ from $\pi_\theta$ for $(x_r,y_r) \in B_r$ via top-$p$ logit filtering
    \State Compute forget loss $\mathcal{L}_{\text{DiPO-f}}$ on $B_f$ using $\pi_w=\pi_f, \pi_l=\pi_m$ \Comment{Based on Eq. \ref{eq:dipo_loss_f}}
    \State Compute retain loss $\mathcal{L}_{\text{DiPO-r}}$ on $B_r$ using $\pi_w=\pi_m, \pi_l=\pi_f$ \Comment{Based on Eq. \ref{eq:dipo_loss_r}}
    \State Compute total loss $\mathcal{L}(\theta) = \mathcal{L}_{\text{DiPO-f}} + \lambda \mathcal{L}_{\text{DiPO-r}}$ \Comment{Using Eq. \ref{eq:final_combined_loss}}
    \State Update parameters $\theta \leftarrow \theta - \eta \nabla_\theta \mathcal{L}(\theta)$
\EndFor
\State \textbf{Output:} Unlearned policy model $\pi_\theta$
\end{algorithmic}
\end{algorithm}

\section{Baseline Methods}
\label{app:baseline}
This section details the baseline methods used for comparison in our experiments. We categorize them into optimization-based methods, which are the primary focus of comparison for our DiPO method, and other unlearning frameworks represented by a state-of-the-art method.
\subsection{Optimization-based method}
Optimization-based methods directly modify the model parameters by minimizing a combined objective function, typically structured as $\mathcal{L}(\theta) = \mathcal{L}_r(\theta) + \lambda\mathcal{L}_f(\theta)$, where $\mathcal{L}_f$ promotes forgetting and $\mathcal{L}_r$ encourages retention, balanced by $\lambda$. We describe common choices for these loss components below.

\subsubsection{Forget losses}
\paragraph{Gradient ascent loss} $\mathcal{L}_{\text{GA}}$ is a fundamental and intuitive unlearning loss function \citep{thudi2022unrolling, maini2024tofu} that aims to maximize the next-token prediction loss on the forget set $\mathcal{D}_f$, which is equivalent to minimizing the likelihood of correct predictions. We denote this forget loss as:
\begin{equation}
    \mathcal{L}_{\text{GA}}(\theta)=\mathbb{E}_{(x_f,y_f)\sim\mathcal{D}_f}[\log \pi_\theta(y_f|x_f)].
    \label{eq:ga-loss}
\end{equation}
While intuitive, $\mathcal{L}_{\text{GA}}$ is unbounded below (likelihood can approach zero), which can lead to training instability and model degradation.
\paragraph{Direct preference optimization loss} $\mathcal{L}_{{\text{DPO}}}$ adapts the Direct Preference Optimization framework \citep{rafailov2023direct} for unlearning \citep{zhang2024negative} (distinguish from standard DPO). It requires a dataset of simple, template-based alternative responses $\mathcal{D}_{a}$ (e.g. $y_{idk}=$ ``I don't know'') and formulates the forget loss to prefer $y_{idk}$ over the original forget response $y_f$:
\begin{equation}
    \mathcal{L}_{{\text{DPO}}}(\theta) = - \frac{1}{\beta}\mathbb{E}_{(x_f,y_f)\sim\mathcal{D}_f, y_{idk}\sim\mathcal{D}_{a}} [\log\sigma(\beta \frac{\pi_\theta(y_{idk}|x_f)}{\pi_{\text{ref}}(y_{idk}|x_f)} - \beta \frac{\pi_\theta(y_f|x_f)}{\pi_{\text{ref}}(y_f|x_f)})].
    \label{eq:app-dpo-template}
\end{equation}
where $\sigma(\cdot)$ is the sigmoid function, $\beta$ is a hyper-parameter controlling the preference strength, and $\pi_\text{ref}$ is reference model (often the initial model before unlearning). This loss is bounded but can suffer from catastrophic forgetting, excessively favoring $y_{idk}$ even for retain queries.
\paragraph{Negative preference optimization loss} $\mathcal{L}_{{\text{NPO}}}$ is a variant of $\mathcal{L}_{{\text{DPO}}}$ for unlearning in recent work \citep{zhang2024negative}. NPO focuses solely on penalizing the forget responses $y_f$ by treating them as dispreferred, without requiring preferred alternatives $y_{idk}$. Its forget loss term is:
\begin{equation}
    \mathcal{L}_{{\text{NPO}}}(\theta) = - \frac{2}{\beta}\mathbb{E}_{(x_f,y_f)\sim\mathcal{D}_f} [\log\sigma(- \beta \frac{\pi_\theta(y_f|x_f)}{\pi_{\text{ref}}(y_f|x_f)})],
\end{equation}
NPO avoids the unboundedness of $\mathcal{L}_{\text{GA}}$ and the need for $y_{idk}$ in $\mathcal{L}_{{\text{DPO}}}$, but lacks an explicit positive preference signal.

\subsubsection{Retain losses}
\paragraph{Gradient descent loss} $\mathcal{L}_{\text{GD}}$ is the standard negative log-likelihood loss applied to the retain set $\mathcal{D}_r$ \citep{maini2024tofu, zhang2024negative}, encouraging the model to maintain its predictive performance:
\begin{equation}
    \mathcal{L}_{\text{GD}}(\theta)=\mathbb{E}_{(x_r,y_r)\sim\mathcal{D}_r}[-\log \pi_\theta(y_r|x_r)]. \label{eq:gd-loss}
\end{equation}
The combination of $\mathcal{L}_{\text{GA}}$ as $\mathcal{L}_{f}$ and $\mathcal{L}_{\text{GD}}$ as $\mathcal{L}_{r}$ constitutes the \emph{GradDiff} method \citep{liu2022continual, yao2024large, maini2024tofu}.
\paragraph{KL-divergence loss} $\mathcal{L}_{\text{KL}}$ aims to preserve the model's behavior by minimizing the KL divergence between the current model $\pi_\theta$ and reference model $\pi_{\text{ref}}$ over the retain set \citep{maini2024tofu, zhang2024negative}:
\begin{equation}
    \mathcal{L}_{\text{KL}}(\theta)=\mathbb{E}_{(x_r,y_r)\sim\mathcal{D}_r}[D_{KL}(\pi_\theta(\cdot|x_r)||\pi_{\text{ref}}(\cdot|x_r))].
\end{equation}

\subsection{Other Unlearning Framework}
\label{app:other_unlearn_frameworks}
Beyond optimization-based fine-tuning, alternative unlearning paradigms exist that employ different mechanisms, such as auxiliary models, data manipulation techniques (see \Cref{s:related-work}). To provide context against strong baselines from distinct research directions within these paradigms, we include two representative methods: \textbf{ULD} \citep{ji2024reversing} and \textbf{AltPO} \citep{mekala2024alternate}. ULD exemplifies methods that achieve unlearning without direct fine-tuning of the target model's parameters, instead relying on an auxiliary model and logit manipulation, representing a strong baseline for \emph{non-optimization-based} unlearning frameworks. AltPO, on the other hand, showcases a \emph{hybrid} approach combined DPO-style losses with data-based techniques.
\paragraph{ULD} This method trains an auxiliary LLM on \emph{augmented versions} of the forget and retain sets ($\mathcal{D}_f^{\prime}$ and $\mathcal{D}_r^{\prime}$, respectively) to perform the \emph{inverse} unlearning task. Specifically, the auxiliary model is trained to maximize likelihood on $\mathcal{D}_f^{\prime}$ (memorizing) while driving its output distribution towards uniform on $\mathcal{D}_r^{\prime}$ (forgetting). The final unlearned model's logits are obtained by subtracting the auxiliary model's logits from the original target model's logits. This approach differs significantly from fine-tuning methods and is particularly noted for its effectiveness in preserving model utility while achieving strong unlearning performance, thus offering a valuable comparison point from a distinct unlearning strategy.
\paragraph{AltPO} This method also employs an auxiliary model, guided by carefully designed prompts, to generate a privacy-preserving alternative response $y_{f_a}$ for each sample in the forget set $\mathcal{D}_f$. This $y_{f_a}$ then replaces the template-based response $y_{idk}$ in \Cref{eq:app-dpo-template}, mitigating catastrophic forgetting. Following its original paper \citep{mekala2024alternate}, the forget loss is denoted as:
\begin{equation}
    \mathcal{L}_{{\text{AltPO}}}(\theta) = - \frac{2}{\beta}\mathbb{E}_{(x_f,y_f)\sim\mathcal{D}_f, y_{f_a}\sim\mathcal{D}_{a}} [\log\sigma(\beta \frac{\pi_\theta(y_{f_a}|x_f)}{\pi_{\text{ref}}(y_{f_a}|x_f)} - \beta \frac{\pi_\theta(y_f|x_f)}{\pi_{\text{ref}}(y_f|x_f)})].
\end{equation}
Similarly, AltPO utilizes $\mathcal{L}_{\text{GD}}$ as its retain loss. Due to the use of an auxiliary model to obtain alternative responses and thereby augment the dataset, it is not classified as a \emph{purely optimization-based method} but rather as a hybrid approach combined with data-based techniques. We include it for comparison against our method, viewing it as \emph{a more advanced development compared to NPO}, particularly in its provision of an explicit, generated positive preference.

\section{Experiments Details}
\label{app:experiment}
\subsection{Hardware configuration}
All experiments are conducted on 2 NVIDIA A800-SXM4-80GB GPU cards in a single node. We employ DeepSpeed ZeRO stage-2  for all baselines to compress GPU memory.
A typical experimental run for our main DiPO method on benchmarks like TOFU or MUSE, involving 10 epochs of training with evaluation performed after each epoch, took approximately 1 hour on this hardware setup.
For our main DiPO method, a complete experimental run on a single task within the MUSE or TOFU benchmarks (typically involving 10 epochs of training with evaluation after each epoch) was generally completed within 1 hour on this hardware setup.

\subsection{Details on Filter mechanism}
\label{app:filter}
This appendix clarifies the \emph{top-k filtering} strategy (using rate $p_k$) mentioned in \Cref{s:preference-pairs-loss-funciton}.
This top-k filtering strategy represents a mechanism for manipulating model output logits, previously employed in various generation contexts \citep{li2022contrastive, chuang2023dola, liu2021dexperts}. Similar to its adoption in related unlearning frameworks like ULD \citep{ji2024reversing}, we utilize it here to determine which tokens' original logits $\mathbf{z}_t$ contribute to the memory vector $\mathbf{m}_t$.

In this section, we will provide a more formal definition. Let $S_t \subset V$ be the set of tokens selected by \emph{top-k filtering},  keeping the top $p_k$ tokens of the vocabulary size.
We define a `memory vector' $\mathbf{m}_t$ that isolates the logits corresponding to these high-confidence tokens: $\mathbf{m}_t = \mathbf{z}_t \odot \text{mask}(\mathbf{z}_t, S_t)$, where $\text{mask}(\mathbf{z}_t, S_t)$ is a binary vector selecting tokens in $S_t$. We then construct the memory-enhancing distribution $\pi_m$ and the forgetting-promoting distribution $\pi_f$ by adding or subtracting this memory vector, scaled by a factor $\alpha$:
\begin{align}
    \pi_m(\cdot | x, y^{<t}) = \text{softmax}(\mathbf{z}_t + \alpha \mathbf{m}_t), \quad
    \pi_f(\cdot | x, y^{<t}) = \text{softmax}(\mathbf{z}_t - \alpha \mathbf{m}_t) .
\end{align}
To determine the set $S_t$, we first compute log-probabilities $\mathbf{s}_t = \text{log\_softmax}(\mathbf{z}_t)$. A dynamic threshold $\tau$ is then established by considering two criteria:

\begin{enumerate}
    \item \textbf{Rank-based Threshold ($\tau_k$):} This ensures at least a minimum number of tokens are kept. It is set to the log-probability corresponding to the $k$-th rank when tokens are sorted by log-probability in descending order, where $k = \max(1, \lfloor p_k \cdot |V| \rfloor)$.
    \item \textbf{Relative Threshold ($\tau_{rel}$):} This adapts to the sharpness of the distribution and is calculated relative to the maximum log-probability: $\tau_{rel} = \max(\mathbf{s}_t) + \log(p_k)$.
\end{enumerate}

The final threshold used for filtering is the minimum of these two: $\tau = \min(\tau_k, \tau_{rel})$. The set $S_t$ then comprises all tokens whose log-probability is greater than or equal to this final threshold ($S_t = \{i \mid s_{t,i} \ge \tau\}$). This ensures that only the logits of these high-confidence tokens are isolated in the memory vector $\mathbf{m}_t = \mathbf{z}_t \odot \text{mask}(\mathbf{z}_t, S_t)$. In this paper, we set $p_k=0.05$.

\subsection{Implementation Details on TOFU}
\label{app:tofu}
\subsubsection{Descriptions of the dataset}
TOFU focuses on unlearning the knowledge of fictitious authors. It contains 200 fictitious author profiles, each consisting of 20 question-answer pairs generated by GPT-4 based on some predefined attributes. These profiles are fictitious and do not exist in the pre-training data, providing a controlled environment for studying unlearning LLMs. TOFU contains three \emph{Forget} set $\mathcal{D}_f$ configurations, each with 1\%, 5\%, and 10\% of the fictional authors, referred to as \text{TOFU-1\%}, \text{TOFU-5\%}, and \text{TOFU-10\%}, respectively. The remaining data constitutes the \emph{Retain} set $\mathcal{D}_r$, used to assess the model's preservation of non-targeted knowledge after unlearning. To further examine unlearning's impact on overall capabilities, TOFU includes two additional evaluation subsets: the \emph{Real Authors} set $\mathcal{D}_{RA}$, for performance on real-world information conceptually related to $\mathcal{D}_f$ but not part of fine-tuning, and the \emph{World Facts} set $\mathcal{D}_{WF}$, for assessing general world knowledge.

\begin{table}[h]
  \caption{Data statistics of Forget set $\mathcal{D}_f$, Retain set $\mathcal{D}_r$, Real Authors set $\mathcal{D}_{RA}$ and World Facts set $\mathcal{D}_{WF}$.}
  \label{tofu data statistics}
  \centering
  \begin{tabular}{ccccc}
    \toprule
    \textbf{Task}     &  $\mathcal{D}_f$   & $\mathcal{D}_r$ & $\mathcal{D}_{RA}$ & $\mathcal{D}_{WF}$  \\
    \midrule
    TOFU-1\%     & 40  & 400  & 100 & 117   \\ 
    TOFU-5\%     & 200 & 400  & 100 & 117   \\
    TOFU-10\%    & 400 & 400  & 100 & 117   \\
    \bottomrule
  \end{tabular}
\end{table}

\subsubsection{Evaluation Metrics}
Our evaluation centers on two primary metrics in the original TOFU paper \citep{maini2024tofu}: Model Utility (MU) and Forget Quality (FQ). 
\paragraph{Model Utility (MU)} This metric quantifies the side effects of unlearning on the model's general knowledge and capabilities. It aggregates performance on the Retain, Real Authors, and Real World sets, considering answer generation probability, ROUGE-L similarity, and Truth Ratio. The Truth Ratio $R_{\text{truth}}$ assesses the model's ability to distinguish factual information, defined as the propensity to generate a paraphrased correct answer ($\tilde{a}$) versus a set of structurally similar but incorrect perturbed answers ($\hat{a}_i$) for a given question ($q$):
\begin{equation*}
R_{truth} := \frac{\frac{1}{5}\sum_{i=1}^{5}\mathbb{P}(\widehat{a}_i|q)^{1/|\widehat{a}_i|}}{\mathbb{P}(\tilde{a}|q)^{1/|\tilde{a}|}} .
\end{equation*}
Here, $q$ is the input question, $P(\cdot|q)$ is the model's probability for a specific answer, $|\cdot|$ denotes answer length in tokens, and $N$ is the number of perturbed answers. MU is the harmonic mean of these three sub-metrics across the three evaluation datasets (nine scores total), a method sensitive to any single low score.

\paragraph{Forget Quality (FQ)} This metric evaluates the success of erasing targeted information $\mathcal{D}_f$. It compares the unlearned model's behavior to that of an ideal reference model (typically trained only on $\mathcal{D}_r$ and thus unexposed to $\mathcal{D}_f$) when queried about $\mathcal{D}_f$. The assessment uses a two-sample Kolmogorov-Smirnov (KS) test on the Truth Ratio distributions from these two models on $\mathcal{D}_f$. A high p-value (e.g. >0.05) indicates no significant distributional difference, suggesting effective unlearning.


\subsubsection{Hyperparameter Implementation} 
Following the setup of \citep{maini2024tofu}, We use the fine-tuned \textbf{LLama2-chat-7B} released by TOFU as the original LLM  and fine-tune the target LLM for 10 epochs. 
For all baseline methods and ours, we set the batch size and learning rate to 32 and $1e-5$ following previous works. We set $\beta$ in \Cref{eq:dipo_loss_f} ($\beta_f$) and \Cref{eq:dipo_loss_r} ($\beta_r$) to 0.05 in our method. For all baseline methods involving retain loss, we set the weight $\lambda$ to 1. More details are in \Cref{app:tofu}.

\subsection{Implementation Details on MUSE}
\label{app:muse}
\subsubsection{Descriptions of the dataset}
MUSE proposes a multi-faceted framework considering six desirable properties, catering to both data owner and model deployer expectations. In this paper, we focus on the News corpus. For this corpus, distinct \textbf{Forget Sets ($\mathcal{D}_\text{forget}$)}, \textbf{Retain Sets ($\mathcal{D}_\text{retain}$)}, and disjoint hold-out sets ($\mathcal{D}_\text{holdout}$) are established as disjoint collections of news articles.
To facilitate granular evaluation, two types of data are derived from these news articles:
\begin{enumerate}
    \item \textbf{Verbatim text}: Original text excerpts from news articles used to assess the prevention of verbatim memorization.
    \item \textbf{Knowledge set}: Question-answer (QA) pairs derived from the original news texts to evaluate the removal of factual knowledge.
\end{enumerate}

\subsubsection{Evaluation Metrics}
MUSE evaluates unlearning across six criteria. We highlight key metrics reflecting data owner and deployer concerns as applied to the NEWS corpus:

\subsubsection*{Data Owner Focused Metrics}
\begin{enumerate}
    \item \textbf{No Verbatim Memorization (VerbMem-f)}: Assesses if the unlearned model ($f_{\text{unlearn}}$) avoids reproducing exact text sequences from the $\mathcal{D}_\text{forget}$ of news articles. Quantified by \emph{VerbMem-f}, which measures the ROUGE-L F1 score between model-generated continuations and true continuations from $\mathcal{D}_\text{forget}$.
    \begin{equation*}
        \text{VerbMem-f}(f, \mathcal{D}_\text{forget}) := \frac{1}{|\mathcal{D}_\text{forget}|} \sum_{x \in \mathcal{D}_\text{forget}} \text{ROUGE-L}(f(x_{[:l]}), x_{[l+1:]}).
    \end{equation*}
    \item \textbf{No Knowledge Memorization (KnowMem-f)}: Measures if $f_{\text{unlearn}}$ can no longer answer questions whose answers are exclusively found in the $\mathcal{D}_\text{forget}$ of news articles. Quantified by \emph{KnowMem-f}, averaging ROUGE scores between model answers and ground-truth answers for QA pairs derived from $\mathcal{D}_\text{forget}$.
    \item \textbf{No Privacy Leakage (PrivLeak)}: Evaluates if the inclusion of news articles from $\mathcal{D}_\text{forget}$ in the original training data ($\mathcal{D}_\text{train}$) can be inferred from $f_{\text{unlearn}}$. Measured by \emph{PrivLeak}, which compares the Area Under the ROC Curve (AUC) of a Membership Inference Attack (MIA) on $f_{\text{unlearn}}$ against that on a perfectly retrained model ($f_{\text{retrain}}$), discriminating between $\mathcal{D}_\text{forget}$ (member news articles) and $\mathcal{D}_\text{holdout}$ (non-member news articles).
    \begin{equation*}
        \text{PrivLeak} := \frac{\text{AUC}(f_{\text{unlearn}}; \mathcal{D}_\text{forget}, \mathcal{D}_\text{holdout}) - \text{AUC}(f_{\text{retrain}}; \mathcal{D}_\text{forget}, \mathcal{D}_\text{holdout})}{\text{AUC}(f_{\text{retrain}}; \mathcal{D}_\text{forget}, \mathcal{D}_\text{holdout})}.
    \end{equation*}
    A \emph{PrivLeak} score close to zero is desirable.
\end{enumerate}

\subsubsection*{Deployer Focused Metrics}
\begin{enumerate}
    \item \textbf{Utility Preservation (KnowMem-r)}: Quantifies how well $f_{\text{unlearn}}$ maintains its performance on the $\mathcal{D}_\text{retain}$ of news articles. This is typically measured using the \emph{KnowMem-r} metric applied to $\mathcal{D}_\text{retain}$: $\text{KnowMem-r}(f_{\text{unlearn}}, \mathcal{D}_\text{retain})$.
    \item \textbf{Scalability}: Assesses how unlearning methods perform with increasing sizes of $\mathcal{D}_\text{forget}$ within the NEWS corpus.
    \item \textbf{Sustainability}: Evaluates performance under sequential unlearning requests involving different sets of news articles.
\end{enumerate}

\subsubsection{Hyperparameter Implementation}
Following the setup of MUSE \citep{shi2025muse}, we use LLaMA-2 7B as the original model, which was released before the collected BBC news articles to prevent potential data leakage.
For baseline methods, we set the batch size to 32, and fine-tune for 5 epochs using AdamW optimizer with a constant learning rate of $1e-5$, For our method, we use the same training hyper-parameters as described in TOFU.

\subsection{Additional Results on TOFU}
\label{app:add-tofu}

\subsubsection{Details on Ablation Study}
\label{app:ablation-studies}
To determine the optimal configuration for our DiPO method, we conducte an ablation study comparing different retain loss functions when combined with the DiPO forget loss component, $\mathcal{L}_{{\text{DiPO-f}}}(\theta)$ (defined in \Cref{eq:dipo_loss_f}). We evaluate the following configurations on the TOFU-10\% task at the best-epoch:
\begin{enumerate}
    \item \textbf{DiPO (ours):} This is the configuration presented as our main result in the paper, using the $\mathcal{L}_{{\text{DiPO-r}}}$ by reversing the roles of the preference distributions of $\mathcal{L}_{{\text{DiPO-f}}}$ on the retain set. The combined objective is then expressed as $\mathcal{L} = \mathcal{L}_{\text{DiPO-f}}(\theta) + \lambda \mathcal{L}_{\text{DiPO-r}}(\theta)$.

    \item \textbf{DiPO(f)+GD:} This configuration utilizes the standard Gradient Descent loss \Cref{eq:gd-loss} on the retain set:
    \begin{align*}
    \min_{\theta} \mathcal{L}(\theta) &= \min_{\theta} \left( \mathcal{L}_{{\text{DiPO-f}}}(\theta) + \gamma \mathcal{L}_{{\text{GD}}}(\theta) \right) \\
    &= \min_{\theta} \left( \mathcal{L}_{{\text{DiPO-f}}}(\theta) + \lambda \mathbb{E}_{(x_r,y_r)\sim\mathcal{D}_r}[-\log \pi_\theta(y_r|x_r)] \right).
    \end{align*}

    \item \textbf{GA+DiPO(r):} This configuration utilizes the standard Gradient Descent loss \Cref{eq:gd-loss} on the retain set:
    \begin{align*}
    \min_{\theta} \mathcal{L}(\theta) &= \min_{\theta} \left( \mathcal{L}_{{\text{GA}}}(\theta) + \lambda \mathcal{L}_{{\text{DiPO-r}}}(\theta) \right) \\
    &= \min_{\theta} \left(\mathbb{E}_{(x_f,y_f)\sim\mathcal{D}_f}[\log \pi_\theta(y_f|x_f)]+ \lambda  \mathcal{L}_{{\text{DiPO-r}}}(\theta) \right).
    \end{align*}

    \item \textbf{NPO+DiPO(r):} This configuration utilizes the standard Gradient Descent loss \Cref{eq:gd-loss} on the retain set:
    \begin{align*}
    \min_{\theta} \mathcal{L}(\theta) &= \min_{\theta} \left( \mathcal{L}_{{\text{NPO}}}(\theta) + \lambda \mathcal{L}_{{\text{DiPO-r}}}(\theta) \right) \\
    \end{align*}

\end{enumerate}

\begin{figure}[h]
    \centering
    \includegraphics[width=0.8\textwidth]{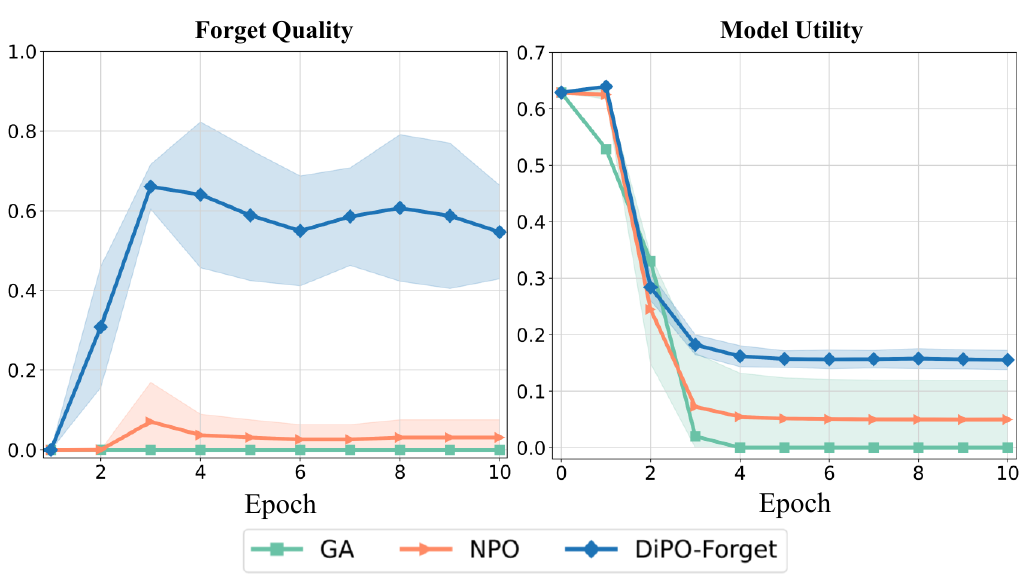}
    \caption{Training curves for only-forget configuration on TOFU-10\%, with GA and NPO curves additionally included for comparison.}
    \label{figure:ablation-curves}
\end{figure}
For these settings, the final results are presented in \Cref{tab:tofu_results_ablation}. Additionally, we discuss DiPO-Forget (using only $\mathcal{L}_{\text{DiPO-f}}$ without any retain loss). This setup simulates scenarios where retain data might be unavailable. we set learning rate to  $7e-6$ and $\beta$ to 0.5 in this configuration. As illustrated by its training dynamics on TOFU-10\% (\Cref{figure:ablation-curves}), even without an explicit retain loss, DiPO-Forget achieves substantial unlearning (e.g. FQ reaching approximately 0.51) while maintaining a notable degree of model utility (e.g. MU around 0.18 at the end of training, after an initial drop). This contrasts sharply with typical baselines where removing the retain loss often leads to a near-complete collapse in both MU and FQ. The ability of DiPO-Forget to preserve some utility while effectively unlearning underscores the inherent stability and targeted nature of the DiPO forget mechanism. This finding is particularly promising for unlearning scenarios where access to comprehensive retain data is limited or unavailable.


\subsubsection{Results at the Final Epoch}
\label{app:tofu-final-epoch}

\begin{table}[h]
\centering
\caption{The final-epoch performance averaged over five seeds on TOFU benchmark. Scores closer to ``Retrain'' are better. \textbf{Bold} indicates best results among all methods.}
\label{tab:tofu_results_final}
\setlength{\tabcolsep}{3pt} 
\begin{tabular}{@{}c|cc|cc|cc@{}}
\toprule
\textbf{Method} & \multicolumn{2}{c|}{\textbf{TOFU-1\%}} & \multicolumn{2}{c|}{\textbf{TOFU-5\%}} & \multicolumn{2}{c}{\textbf{TOFU-10\%}} \\
\cmidrule(lr){2-3} \cmidrule(lr){4-5} \cmidrule(lr){6-7}
 & FQ $\uparrow$ & MU $\uparrow$ & FQ $\uparrow$ & MU $\uparrow$ & FQ $\uparrow$ & MU $\uparrow$ \\
\midrule
Original LLM & 1e-3 & 0.62 & 3e-16 & 0.62 & 2e-19 & 0.62 \\
Retrain LLM & 1.0 & 0.62 & 1.0 & 0.62 & 1.0 & 0.62 \\
\midrule
GA & 0.40 & 0.52 & 5e-8 & 0 & 6e-11 & 0 \\
GA+GD & 0.27 & 0.53 & 0.11 & 0.33 & 9e-3 & 0.51 \\
GA+KL & 0.31 & 0.53 & 0.14 & 0.35 & 1e-5 & 0.55 \\
\midrule
NPO  & 0.71 & 0.56 & 0.03 & 0.02 & 5e-4 & 0 \\
DPO+GD & 0.27 & \textbf{0.58} & 1e-4 & 0.02 & 5e-7 & 0 \\
NPO+GD & 0.73 & \textbf{0.58} & 0.64 & 0.57 & 0.17 & 0.53 \\
\midrule
DiPO & \textbf{0.89} & \textbf{0.58} & \textbf{0.95} & \textbf{0.58} & \textbf{0.84} & \textbf{0.56} \\
\bottomrule
\end{tabular}
\end{table}

\end{document}